\documentclass[10pt,journal,compsoc]{IEEEtran}
\usepackage{soul}
\usepackage{url}
\usepackage[utf8]{inputenc}
\usepackage[small]{caption}
\usepackage{graphicx}
\usepackage{amsmath}
\usepackage{amsthm}
\usepackage{booktabs}
\usepackage{xcolor}
\usepackage{algorithmic}
\usepackage[ruled]{algorithm2e}
\usepackage{comment}
\usepackage{subfigure}
\usepackage{float}
\usepackage{bbm}
\usepackage{multirow} 

\newcommand{\RNum}[1]{\uppercase\expandafter{\romannumeral #1\relax}}
\newtheorem{Def}{Definition}
\newtheorem{Theo}{Theorem}

\newtheorem{lem}{Lemma}

\ifCLASSINFOpdf
\else
\fi

\begin{document}
%
\title{Fairness in Semi-supervised Learning: \\ Unlabeled Data Help to Reduce Discrimination}
%
%
%

\author{Tao Zhang,
	Tianqing Zhu, Jing Li, Mengde Han, Wanlei Zhou, ~\IEEEmembership{Senior Member, IEEE} and Philip S. Yu~\IEEEmembership{Fellow, IEEE}

\thanks{ 
	Tianqing Zhu$ ^{*} $ is the corresponding author.
	
	Tao Zhang, Tianqing Zhu,  Mengde Han, Wanlei Zhou are in centre for cyber security and privacy with the School of Computer Science, University of Technology, Sydney,
	Australia. Email: $ \{ $Tao.Zhang-3@student.uts.edu.au,  Tianqing.Zhu@uts.edu.au, Mengde.Han@student.uts.edu.au, Wanlei.Zhou@uts.edu.au$ \} $
	
	Jing Li is in the centre for Artificial Intelligence, University of Technology Sydney. Email: $ \{ $jing.li-20@student.uts.edu.au$ \} $
	
	Philip S, Yu is with the Department of Computer Science University of Illinois at Chicago Chicago, Illinois, USA. 
	Email: $ \{ $psyu@uic.edu$ \} $

}}
%
%

\markboth{Journal of \LaTeX\ Class Files,~Vol.~14, No.~8, August~2015}%
{Shell \MakeLowercase{\textit{et al.}}: Bare Demo of IEEEtran.cls for IEEE Journals}
%




\IEEEtitleabstractindextext{%
\begin{abstract}
	A growing specter in the rise of machine learning is whether the decisions made by machine learning models are fair. While research is already underway to formalize a machine-learning concept of fairness and to design frameworks for building fair models with sacrifice in accuracy, most are geared toward either supervised or unsupervised learning. 
	Yet two observations inspired us to wonder whether semi-supervised learning might be useful to solve discrimination problems. First, previous study showed that increasing the size of the training set may lead to a better trade-off between fairness and accuracy. Second, the most powerful models today require an enormous of data to train which, in practical terms, is likely possible from a combination of labeled and unlabeled data. 
	Hence, in this paper, we present a framework of fair semi-supervised learning in the pre-processing phase, including pseudo labeling to predict labels for unlabeled data,  a re-sampling method to obtain multiple fair datasets and lastly, ensemble learning to improve accuracy and decrease discrimination.
	A theoretical decomposition analysis of bias, variance and noise highlights the different sources of discrimination and the impact they have on fairness in semi-supervised learning. 
	A set of experiments on real-world and synthetic datasets show that our method is able to use unlabeled data to achieve a better trade-off between accuracy and discrimination.
	
\end{abstract}

}

\maketitle
\IEEEdisplaynontitleabstractindextext

%
\IEEEpeerreviewmaketitle

\IEEEraisesectionheading{\section{Introduction}}
Machine learning is now in wide use as a decision-making tool in many areas, such as job employment, risk assessment, loan approvals and many other basic precursors to equity. However, the popularity of machine learning has raised concerns about whether the decisions algorithms make are fair to all individuals.
For example,  Chouldechova found evidence of racial bias in recidivism prediction tool where black defendants are more likely to be assessed with high risk than white defendants \cite{chouldechova2017fair} . Obermeyer et al. found prejudice in health care systems where black patients assigned the same level of risk by the algorithm are sicker than white patients \cite{Obermeyer447}.  These findings show that unfair machine learning algorithms will affect legal justices, health care, and other aspects of human beings.

As we move forward in a world of machine-assisted predictions for human-beings, the fairness of machine learning has become a very cardinal issue. In the future, our ability to design machine learning algorithms that treat all groups equally may be one of the most influential factors in who will be the haves and who will be the have-nots.
As the influence and scope of these risk assessments increase, academics, policymakers, and journalists have raised concerns that the statistical models from which they are derived might inadvertently encode human biases

Over the past few years, much research has been devoted to designing fairness metrics, such as statistical fairness \cite{chouldechova2017fair,calders2009building,pmlr-v54-zafar17a,hardt2016equality}, individual fairness \cite{dwork2012fairness,louizos2015variational,jung2019eliciting} and causal fairness \cite{kusner2017counterfactual,kilbertus2017avoiding}. 
These approaches and algorithms can be roughly divided into three categories: pre-processing methods, in-processing methods and post-processing methods.
Pre-processing methods adjust data distribution \cite{calders2009building,kamiran2012data} or learn new fair representations \cite{zemel2013learning,madras2018learning,pmlr-v89-song19a}, to relieve some of the tension between accuracy and fairness. In-processing methods add constraints or regularizers  to restrict the correlation between labels and sensitive/protected attributes, i.e., traits that can be targets for discrimination \cite{pmlr-v54-zafar17a,kamishima2012fairness,pmlr-v97-kleindessner19b}. 
Post-process methods calibrate training results \cite{hardt2016equality}. 
These studies mainly focus on addressing the two most crucial fundamental issues in machine learning fairness: how to formalize the concept of fairness in the context of machine learning tasks, and how to design effective algorithms to achieve an ideal compromise between accuracy and fairness.


However, almost all methods achieving fairness are mostly for either supervised learning or unsupervised learning, and fair semi-supervised learning (SSL) has rarely been considered. Realistically though, training data is often a combination of labeled and unlabeled samples, so a semi-supervised solution has high practical value. 
Also, since “ideal” is a lofty goal, the trade-off between accuracy and fairness is still an ongoing pursuit. \cite{NIPS2018_7613} showed that increasing the amount of training data is likely to produce a better trade-off between accuracy and fairness. This insight inspired us to wonder whether using unlabeled data to augment the training set might give us a kind of control value with which to balance fairness and accuracy. Unlabeled data is abundant and, if it could be used as training data, we could adjust the size of the training set as required to meet accuracy vs. fairness thresholds.
We may even be able to avoid the need to make a compromise between fairness and accuracy entirely. This leaves fair semi-supervised learning with two challenges: 1) How to make use of unlabeled data to achieve a better trade-off between accuracy and fairness; and 2) How to alleviate the impact of noise, which is common to semi-supervised learning.

To tackle these challenges, we propose a framework to achieve fair SSL in the pre-processing phase. 
The solution to the trade-off challenge is to use unlabeled data to reduce representation discrimination. (Representation discrimination is due to certain parts of the input space under-represented.)
Therefore, the first two steps in our framework are pseudo labeling and re-sampling. The first step is to use pseudo labeling as a SSL method to predict labels for unlabeled data. The second step involves dividing the dataset into groups based on the protected attribute and the label, and then obtain fair datasets by re-sampling the same number of data points in each group.
When unlabeled data is used as training data, it is likely to obtain more under-represented data points from unlabeled data to reduce representation discrimination, and thus to make little compromise between fairness and accuracy. 
The issue of noise induced by (incorrectly) predicting labels for unlabeled data is addressed by the third step in the framework: ensemble learning. Predicting unlabeled data will induce some noise in the labels of unlabeled data. Ensemble learning helps to reduce label noise and the variance of the training model, and to produce more accurate final predictions.

In summary, the contributions of this paper are listed as below.
\begin{itemize}

	\item First, we use unlabeled data to reduce representation discrimination, and thus achieve a better trade-off between accuracy and discrimination.
	
	\item Second, we propose a fairness-enhanced sampling (FS) framework that combines pseudo labeling, re-sampling and ensemble learning for fair SSL in the pre-processing phase. 
	\item Third, we theoretically analyze the sources of discrimination in SSL via bias, variance and noise decomposition, and conduct experiments with both real and synthetic data to validate the effectiveness of our proposed FS framework.
\end{itemize}
The rest of this paper is organized as follows. The background is presented in Section \RNum{2}, and the proposed FS framework is given in Section \RNum{3}. Section \RNum{4} presents the discrimination analysis, and the experiments are set out in Section \RNum{5}. The related work appears in Section \RNum{6}, with the conclusion in Section \RNum{7}.

\section{Background}
\subsection{Notations}
For simplicity, let $ \mathcal{D}_{l} = \{X,A,Y_{l}\}  $ be a dataset with $ N_1 $ data points, where $ X = (X_1, X_2,..., X_d) $ denotes $ d $ unprotected attributes; $ A $ denotes protected attributes, e.g., gender or race; and $ Y_{l} \in \{0,1\}$ is the label for the task. Let $ \mathcal{D}_{u} =  \{X,A,Y_{u} \}  $  be an unlabeled dataset with $ N_{2} $ data points and $ Y_{u} \in \{0,1\} $ be the predicted labeled for the unlabeled dataset.
For ease, assume the protected attribute is binary valued. For example, if the protected attribute is race, the value might be either 'white' $  (A=0)$ or 'black' $ (A=1) $.

Our objective is to learn a mapping $ f(\cdot) $ over a discriminatory dataset $ \mathcal{D}_{l}$ and $\mathcal{D}_{u}$, in which the classification result is independent of protected attributes. 
Performance is measured by both accuracy and the level of discrimination in the results. The ideal classifier should have a high accuracy without discrimination.
\subsection{Fairness Metrics}
Fairness is often evaluated with respect to protected/unprotected groups of individuals defined by attributes, such as gender or age. Here, we have opted for demographic parity as the fairness metrics in this paper.

\begin{Def}
	$\textbf{(Demographic parity)}$ \cite{calders2009building} Demographic parity requires that the probability of a classifier’s prediction be independent of any sensitive attributes, where the probability of the predicted positive labels in group $ a \in \mathcal{A}$ is defined as follows:
\end{Def}
\begin{equation}
\gamma_{1}(\hat{Y}) = Pr(\hat{Y}=1|A=1)
\end{equation}
\begin{equation}
\gamma_{0}(\hat{Y}) = Pr(\hat{Y}=1|A=0)
\end{equation}
\begin{Def}
	$\textbf{(Discrimination level)}$ The discrimination level  $ \gamma $ in terms of demographic parity can be evaluated by the difference between groups,
\end{Def}
\begin{equation}
\Gamma(\hat{Y}) = |\gamma_{0}(\hat{Y})-\gamma_{1}(\hat{Y})|
\end{equation}
\subsection{Discrimination Sources}
Discrimination can exist in every stage of machine learning. Roughly, discrimination sources can be divided into two lines: data discrimination and model discrimination \cite{suresh2019framework}. Our proposed FS method is able to reduce the representation discrimination in the data.
\subsubsection{Data Discrimination}
Data discrimination includes historical discrimination, representation discrimination, measurement discrimination. 
Historical discrimination occurs when there is a discrepancy between the world itself and the values or goals in the model to be encoded and propagated. 
It can stem from cultural stereotypes among people, such as social class, race, nationality, gender.
Representation discrimination occurs when the data used to train the algorithm does not accurately represent the problem space.
As a consequence, the model generalizes to fit the majority groups much than minority groups. 
Measurement discrimination comes from the way we choose, utilize, and measure specific features. The selected set of features and labels may miss important factors, or bring in group or input-related noise that causes different performance.
\subsubsection{Model Discrimination}
Model discrimination includes aggregation discrimination, evaluation  discrimination, deployment discrimination. 
Aggregation discrimination can arise during model construction when different populations are improperly grouped together.
In many applications, the groups of interest are heterogeneous, so a single model is unlikely to fit all subgroups.
Evaluation discrimination occurs during model iteration and evaluation. This can happen when a test or external benchmark unequally represents each group in the population.
Evaluation discrimination may also occur due to the use of performance metrics that are not appropriate for the way the model is used.
Deployment discrimination occurs after the model is deployed when the system is used or interpreted in an inappropriate way.


\subsection{Bias, Variance and Noise}
Following \cite{NIPS2018_7613}, our analysis of discrimination is based on bias, variance and noise decomposition.
First, we present the definition of main prediction. The main prediction for a loss function $ L $ and set of training sets $ D $ is defined as, $y_{m}(x,a)= argmin_{y^{\prime}} \mathbbm{E}_{D}[L(Y,y^{\prime})|X=x,A=a]$, where $ Y $ is the true value; $ y^{\prime} $ is the predicted label with the minimum average loss relative to all the predictions. The expectation is taken with respect to the training sets in $ D $.
\begin{Def}
	(Bias,variance and noise) Following \cite{domingos2000unified},  the bias $ B $, variance $ V $ and noise $ N $ at a point $ (x,a) $ with a model $ f $ are defined as,
	\begin{align}
	&B(f,x,a)=L({y}^{*}(x,a),y_{m}(x,a))\\
	&V(f,x,a)= \mathbbm{E}_{D}[L(y_{m}(x,a),\hat{y}_{D}(x,a)]\\
	&N(f,x,a)=\mathbbm{E}_{Y}[L({y}^{*}(x,a),Y)]
	\end{align}	
\end{Def}	
\noindent	where ${y}^{*}$ is the optimal prediction that achieves the smallest expected error.
Bias is the loss between the main prediction and the optimal prediction. Variance is the average loss incurred by predictions relative to the main prediction from different datasets $ D $.
Noise is the unavoidable component of the loss, which is independent of the learning model.

 Bias, variance and noise decomposition are appropriate tools for analyzing discrimination because loss function relates to the misclassification rate. For example, when using zero-one loss function, the misclassification rate is denoted as
\begin{align}
&\mathbbm{E}[L(y,\hat{y})]=\mathbbm{E}[ \hat y \neq y|a=0] + \mathbbm{E}[\hat y \neq y|a=1]\\ \notag
&=\mathbbm{E}[ \hat{y}=1| y=0,a=0] + \mathbbm{E}[\hat{y}=0 | y=1,a=1] \notag
\end{align}
where $ \hat{y} $ is the predicted label of a classifier. Note that loss function can be decomposited into false positive rate  and false negative rate. And, once false positive rate  and false negative rate are obtained, true positive rate and true negative rate can be obtained. 
As such, many fairness metrics, such as demographic parity and equal opportunity, can be explained by bias, variance and noise decomposition.
\begin{figure*}[htp]
	\centering
	\includegraphics[scale=0.62]{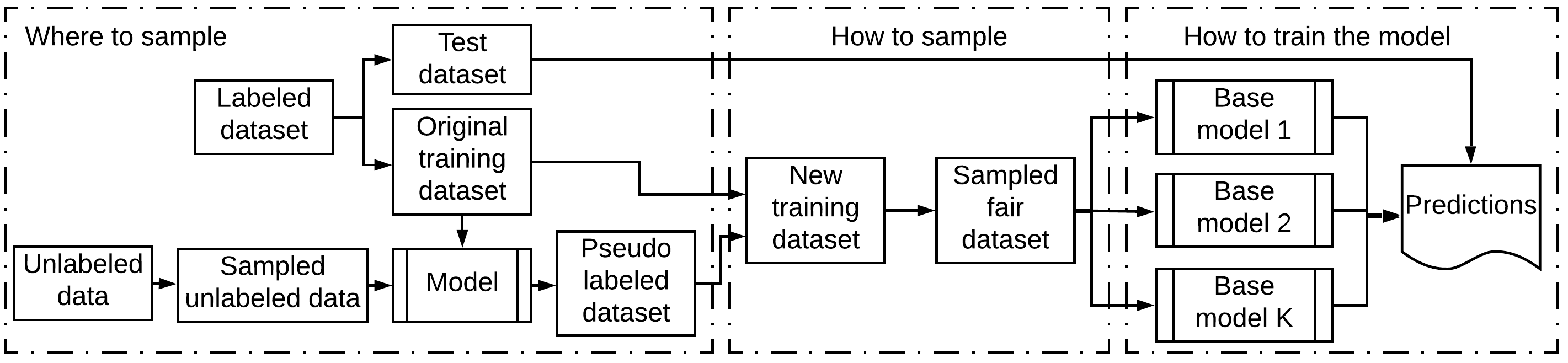}
	\caption{The three phases of the fairness-enhanced sampling framework: 1) where to sample, 2) how to sample and 3) how to train the model.
		Step 1 is to generate a new training dataset which consists of the original dataset and the pseudo labeled dataset. Step 2 is to construct multiple fair datasets through re-sampling. Step 3 is to train a model with each of the fair datasets through ensemble learning to produce the final predictions. }
\end{figure*}
\section{The Proposed Method}
\subsection{Overview of the Fairness-enhanced Sampling Framework}
Figure 1 shows the general description of the fairness-enhanced sampling framework in the pre-precessing phase. The framework consists of three steps: 1) pseudo labeling, 2) re-sampling and 3) fair ensemble learning. The first step is to predict labels for unlabeled data as more data points in the protected group are likely to be found in unlabeled data. The second step is to construct  new datasets that is able to represent all groups equally when the datasets are used for training. In this way, representation discrimination can be removed from training datasets. The third step is to train multiple base models based on multiple fair datasets and final predicted results are obtained from multiple base models. Ensemble learning is able to reduce the label noise that is induced via pseudo labeling, and  the model variance. Each of these steps is discussed in more detail in the following.
\subsection{Where to Sample}
The goal of this step is to use a labeled dataset and part of an unlabeled dataset to construct a new training dataset, as shown in Figure 1.
Suppose we have a labeled dataset $ \mathcal{D}_{l} $ and a large unlabeled dataset $ \mathcal{D}_{u} $.
First, we use the labeled dataset and part of the unlabeled dataset to generate a new training dataset.
With a sample ratio of $ \rho $, we take random samples from the unlabeled dataset $ \mathcal{D}_{u} $ and form sampled unlabeled datasets $ \mathcal{D}_{su} $.
Then we use pseudo labeling to predict the labels for unlabeled data as if they were true labels. Pseudo labeling is a simple and efficient method to implement SSL  \cite{lee2013pseudo}. The procedure, as shown in $ \textbf{Algorithm 1} $, is as follows.

1) Set a split rate $ s \in (0,1) $ and split the labeled dataset into training and test dataset, denoted as the original training dataset and test dataset.
2) Select a learning model and, train the model on the original training dataset to produce a trained model.
3) Use the trained model on $ \mathcal{D}_{su} $ to predict the output (or pseudo label), and the pseudo labeled dataset is obtained. We do not know if these predictions are correct, but we now have predicted labels, which is our goal in this step.
4) Concatenate the original training dataset and pseudo labeled dataset to form a new training dataset $ \mathcal{D}_{new} $. 
\begin{algorithm}[h]
	\caption{Pseudo labeling}
	\LinesNumbered 
	\KwIn{Labled dataset $ \mathcal{D}_{l} $, unlabeled dataset $ \mathcal{D}_{u} $, split rate $s$, sample ratio $ \rho $}
	\KwOut{New training dataset $ \mathcal{D}_{new} $}
	Split $ \mathcal{D}_{l} $ into the training dataset and the test dataset\; 
	Sample  $\mathcal{D}_{su}  $ from  $\mathcal{D}_{u}$\;
	Select a learning model and train the model on the training dataset\;
	Obtain the trained model\;
	Use the trained model to predict labels for $ D_{su}$\;
	Combine the original training dataset and the pseudo labeled dataset to create $ D_{new}$\;
\end{algorithm}

Pseudo-labeling is an easy-to-implement and efficient semi-supervised learning method and, by the above method, can take advantage of unlabeled data to both: a) increase the size of the training set; and b) create more data samples representing minority groups to produce fairer training sets.
Moreover, the learning model can be any models, such as logistic regression, neural networks, etc.

\subsection{How to Sample}
In this step, the goal is to sample multiple fair datasets from the new training datasets to ensure fair learning.  
The rationale for this method is that, since the classifier is trained on non-discriminatory data, its prediction may also be non-discriminatory \cite{kamiran2012data}. 
For simplicity, this analysis covers a binary classification task with one protected attribute, and applies demographic parity as the fairness metric. Our method can certainly be applied to cases with multiple sensitive attributes, subjected to the fairness metrics. 

Based on this setup, the dataset is divided into four groups according to the protected attribute and labeled-values: 1)  Protected group with positive labels ($G_{PP}$), 2)  Unprotected group with positive labels ($G_{UP}$), 3) Protected group with negative labels ($G_{PN}$), and 4)  Unprotected group with negative labels ($G_{UN}$). These divided groups can be denoted as follows,
\begin{align}
& G_{PP} = \{X \in D | A = 1,Y=1\} \\
& G_{UP} = \{X \in D | A = 0,Y=1\} \\
& G_{PN} = \{X \in D | A = 1,Y=0\} \\
& G_{UN} = \{X \in D | A = 0,Y=0\} 
\end{align}
where $ Y=1 $ denotes the positive class and $ Y=0 $ denotes the negative class. $A=1 $ denotes that the data point is in the protected group and  $ A=0 $ denotes that the data point is in the unprotected group. 
To ensure fair learning in the pre-processing phase, the number of data points in the training set for each group should be the same, otherwise the model will fall prey to data discrimination. 
In the case of discrimination, the size of each group is different. Our aim is to adjust the data points by sampling to reach the same size in each group.

$\textbf{Algorithm 2}  $ describes the process of how to obtain multiple fair datasets, and the procedure is as follows:
First, we compute the size of the groups $ G_{PP} $, $ G_{UP} $, $ G_{PN} $, $ G_{UN} $. The sample size is denoted as $ n_{s} $, which means that the number of $ n_{s} $ data points will be sampled from each group. Here, there are two cases:
1) When $ n_{i} \geq n_s $, $ n_s $ data points are sample randomly from the group $ G_{i} $.
2) When $ n_{i} < n_s $, $ n_s $ data points are oversampled from the group $ G_{i} $.
Then we can obtain the fair dataset $ \mathcal{D}_{sf} $ which consists of the number of data points equally for each of the four groups. Repeating this procedure $ K $ times produces $ K $ fair datasets with some commonalities and some differences due to the random sampling, which is desirable for ensemble learning. The next step is to learn from these multiple fair datasets to achieve more accurate and less discriminatory results. 

\begin{algorithm}[h]
	\caption{Fair re-sampling}
	\LinesNumbered 
	\KwIn{ New training datase $ \mathcal{D}_{new} $, sensitive attribute $ A $, sample times $ K $, sample size $ n_{s} $, sample ratio $\rho  $}
	\KwOut{Fair datasets $ \mathcal{D}_{sf} $}
	Divide the dataset into four groups $G_{PP}$, $G_{PN}$, $G_{UP}$, $G_{UN}$ \\
	Calculate the size of all groups $ n_i $ \\
	\For{$ k \in K $}{
		\If{$ n_{i}  \geq n_{s}$ }{
			Sample randomly the number of $ n_s $ data points from the group $ i $
		}
		\eIf{$ n_{i}  \leq n_{s}$}{
			Oversample the number of $ n_{s} $ data points from the group $ i $
		}
		
		Obtain fair datasets $ \mathcal{D}_{sf,i} $
	}
	Obtain multiple fair datasets $ \mathcal D_{sf,1}, \mathcal D_{sf,2}, ..., \mathcal D_{sf,K} $
\end{algorithm}
\subsection{How to Train the Model}
In this step, the goal is to achieve more accurate and less discriminatory training results on multiple fair datasets  $ \mathcal{D}_{sf} $. After obtaining multiple  $ \mathcal{D}_{sf} $, we choose a learning model to train multiple  $ \mathcal{D}_{sf} $ and apply ensemble learning to combine the learning results.
Ensemble learning in machine learning  exploits the independence between base models to improve the overall performance.
In this case, we use Bagging \cite{breiman1996bagging} to 
combine the decisions from multiple base models learned on multiple fair datasets to improve the accuracy and decrease the discrimination.

$ \textbf{Algorithm 3} $ describes the fair ensemble learning. With the new training dataset $ \mathcal D_{new} $ from $\textbf{Algorithm 1} $ and fair datasets  $ \mathcal D_{sf,1}, \mathcal D_{sf,2}, ..., \mathcal D_{sf,K} $ from $\textbf{Algorithm 2} $, train each fair dataset on its own model $ f_{k}(\mathcal D_{sf,k}) $  in parallel. The final model will average the outputs based on the aggregation of predictions from all base models. The predictions obtained from most base models are predicted as final predictions, which is presented as,
\begin{equation}
f(\cdot)= {argmax}_{y \in \mathcal{Y}} \sum_{k=1}^{K} \mathbbm{I} (y=f_k \mathcal(D_{sf,k}   ))
\end{equation}
where $ \mathbbm{I}(\cdot)  $ is the indicator function, and $ K $ is the ensemble size, i.e., the number of fair datasets.

Having some diversity across the datasets is crucial for ensemble learning.
In our approach, the randomness of the fair datasets reflects in two places:
1) randomly sampling the unlabeled dataset $\mathcal{D}_{u} $, and subsequently, the pseudo labeled dataset process in $ \textbf{Algorithm 1} $; and
2) randomly sampling $ n_{s} $ data points for all groups from $\mathcal{D}_{new} $ when constructing each fair dataset.


With ensemble learning, the discrimination level is determined by final predictions.
We redefine the discrimination level in ensemble learning as
$\gamma_{En}= |Pr(f(\cdot)=1|A=1)-Pr(f(\cdot)=1|A=0)|$.
Overall, a combination of multiple base models helps to decrease discrimination resulting from variance and noise, and is able to give a more reliable prediction than a single model. 
\begin{algorithm}[h]
	\caption{Fair ensemble learning}
	\LinesNumbered 
	\KwIn{Dataset, sample times $ K $, sample size $ n_{s} $, split rate $ s $, sample ratio $\rho  $}
	\KwOut{ Accuracy $ Acc $, Discrimination $ \gamma $}
	Execute $ \textbf{Algorithm 1} $ to obtain the new training dataset $\mathcal{D}_{new}$\;
	\For{$ k \in K $}{

		Execute $ \textbf{Algorithm 2} $ to obtain the fair dataset $\mathcal{D}_{sf,k}$\;
		Train the selected model on the fair dataset $\mathcal{D}_{sf,k}  $ and obtain the base model $ f_{k}(\cdot)$\
	}
	
	Make predictions using the final model with ensemble size $ K $ in Eq.(12)\;
\end{algorithm}

\subsection{Discussion}
In reviewing the complete framework, there are several benefits to this approach, which are worth highlighting.
\begin{itemize}
	\item  Many semi-supervised learning methods can be used to predict labels for unlabeled data, such as graph-based learning and transductive support vector machines \cite{zhu2005semi}. We choose pseudo labeling because it is a commonly used semi-supervised learning technique, which is efficient and easy to implement.
	\item The proposed FS framework only removes representation discrimination. However, it is likely that many types of discrimination exist in machine learning, such as historical discrimination, measurement discrimination. Other discrimination can be removed by in-processing or post-processing methods, based on our proposed FS framework. 
\end{itemize}

\section{Discrimination Analysis}
Following \cite{NIPS2018_7613}, we analyze the fairness of the predictive model via bias, variance, and noise decomposition.
The source of discrimination can be decoupled as discrimination in bias $ {B_a} {(f)} $, discrimination in variance $ {V_{a}}{(f)}$ and discrimination in noise $ {N_{a}} $.
The expected discrimination level $ \Gamma(f) $ of a classifier $ f $ learned from a set of training set $ D $ is defined as,
$ \bar{\Gamma}(f)= |\mathbbm{E}_{D}[\Gamma_{0}(f)-\Gamma_{1}(f)]| $. 
\begin{lem}
	The discrimination with regard to group $ a \in A $ is defined as, 
	\begin{equation}
	\gamma_{a}({f}) = \bar{B_a} {(f)} + \bar{V_{a}}{(f)} +  \bar{N_{a}}
	\end{equation}
\end{lem}
Given two groups, the discrimination level is denoted as,
\begin{equation}
\bar{\Gamma} = |(\bar{B_{0}}{(f)}-\bar{B_{1}}{(f)}) + (\bar{V_{0}}{(f)}-\bar{V_{1}}{(f)}) + (\bar{N_{0}}-\bar{N_{1}}) |\\ \notag
\end{equation}
And, in more detail, the discrimination components of Eq.(13), i.e., bias, variance and noise are as follows: 
\begin{align}
&\bar{B_a} ({f}) =\mathbbm{E}_{D} [B(y_{m},x,a)|A=a] \\
&\bar{V_a} ({f}) =\mathbbm{E}_{D} [c_{v}(x,a)V(y_{m},x,a)|A=a] \\
&\bar{N_a} = \mathbbm{E}_{D} [c_{n}(x,a) L(y^{*}(x,a), Y)|A=a] 
\end{align}
where $ c_v(x,a) $ and $ c_{n}(x,a) $ are parameters related to the loss function.
For more details, see the proof in \cite{NIPS2018_7613}.
\begin{lem}
	The discrimination learning curve $ \bar{\Gamma}({f},n) :=|\bar{\gamma_{0}}({f},n) - \bar{\gamma_{1}}({f},n)| $ is asymptotic and behaves as inverse power law curve, where $ n $ is the size of the training data  \cite{NIPS2018_7613}.
\end{lem}
\begin{Theo}
	Unlabeled data is able to reduce discrimination with the proposed FS framework,  if $ (| \bar{V}_{a} ({f})_{sl}|-|\bar{V}_{a} ({f})_{ssl}|)  - \bar{N}_{a,p} \geq 0$.
\end{Theo}
\begin{proof}
	To prove the above theorem, we shall prove that the discrimination level in SSL $\bar{\Gamma}_{ssl}$ is lower than the discrimination level in supervised learning $\bar{\Gamma}_{sl}$. In the following, we will analyze the discrimination in SSL in terms of $\emph{discrimination in bias}$ $\bar{B_a} ({f})_{ssl}$, $\emph{discrimination in variance}$ $\bar{V_a} ({f})_{ssl}$, $ \emph{and discrimination in noise} $ $\bar{N}_{a,ssl}$.
	
	$\emph {Discrimination in Bias} $
	Bias measures the fitting ability of the algorithm itself, and describe accuracy of the model. Hence, bias in discrimination $ \bar{B_a} ({f}) =\mathbbm{E}_{D} [B(y_{m},x,a)|A=a]  $ only depends on the model. When the same model is trained on  the original training dataset and new training dataset, discrimination in bias is the same in supervised learning and SSL, which can be expressed as $ |\bar{B} ({f})_{sl}| - | \bar{B} ({f})_{ssl}| =0 $.
	
	$ \emph{Discrimination in Variance} $
	Discrimination in variance $ \bar{V_a} ({f}) $ can be reduced with extra unlabeled data in the training dataset.
	$ \textbf{Lemma 2} $ states that the discrimination level $ \bar{\Gamma}({f},n) $ decreases with the increasing size of training data $ n $. 
	In our proposed FS framework, unlabeled data is pseudo-labeled, and the new training dataset consists of the original training dataset and the pseudo labeled dataset. 
	The size of the new training dataset can be guaranteed to be larger than the size of the original training by adjusting the sampling size. 
	Also, using Bagging to combine all the base models to obtain the final predictions helps to construct the aggregate model with a lower variance, thus reducing the discrimination in variance $ \bar{B_a} $. 
	Hence, we conclude that $ | \bar{V}_{a} ({f})_{ssl}| -   | \bar{V}_{a} ({f})_{sl}| \leq  0$.
	
	$ \emph{Discrimination in Noise} $
	Unlabeled data introduces more discrimination in noise because pseudo labeling contains discrimination from the trained model. Thus, noisy labels from pseudo labeling in the unprotected group is more than that in the protected group.
	We divide the discrimination in noise in SSL into discrimination in noise in labeled data $\bar{N}_{a,l} $ and discrimination in noise in pseudo labeled data $ \bar{N}_{a,p} $, which is expressed as,
	\begin{equation}
	\bar{N}_{a,ssl} = \bar{N}_{a,l} +\bar{N}_{a,p}
	\end{equation}
	Discrimination in noise in labeled data $ \bar{N}_{a,l} $ is the same as the discrimination in noise in supervised learning $ \bar{N}_{a,sl} $. 
	Then we analyze the discrimination in noise due to pseudo labeled data $ \bar{N}_{a,p} $, including four mislabeled cases,
	\begin{align}
	& \bar{N}_{y=0,a=0}=\mathbbm{E}_{D_{un}}[\hat{y}^*_{p}=1|y=0,a=0] \\
	& \bar{N}_{y=0,a=1}=\mathbbm{E}_{D_{un}}[\hat{y}^*_{p}=1|y=0,a=1] \\
	& \bar{N}_{y=1,a=0}=\mathbbm{E}_{D_{un}}[\hat{y}^*_{p}=0|y=1,a=0] \\
	& \bar{N}_{y=1,a=1}=\mathbbm{E}_{D_{un}}[\hat{y}^*_{p}=0|y=1,a=1] 
	\end{align}
	where $ \hat{y}^*_{p} $ is the optimal predicted label of unlabeled data via pseudo labeling. The noise in the protected group is  $ \bar{N}_{1,p}=\bar{N}_{y=0,a=1} + \bar{N}_{y=1,a=1} $ and the noise in the unprotected group is $ \bar{N}_{0,p}=\bar{N}_{y=0,a=0} + \bar{N}_{y=1,a=0} $. 
	The model contains discrimination because the model is trained on a dataset without any fairness guarantees, and thus the model will bring discrimination in pseudo labeling. In this way, discrimination in noise in pseudo labeled data $ \bar{N}_{a,p} $ can be measured as,
	\begin{align}
	& \bar{N}_{a,p} = |\bar{N}_{1,p} - \bar{N}_{0,p}|
	\end{align}
	To relieve the noise from pseudo labeling, we use Bagging - a robust model that is resilient to class label noise since the errors incurred by the noise can be compensated by the combined predictions of other learners. 
	
	Based on the analysis above, we conclude that when  $ | \bar{V}_{a} (\hat{Y})_{ssl} - \Delta \bar{V}_{a} (\hat{Y})_{sl}|- \bar{N}_{a,p} \geq 0$,  unlabeled data is able to reduce discrimination with the proposed FS framework. 
	Unlabeled data do not change discrimination in bias. However, they do reduce discrimination in variance, and they increase discrimination in noise, but bagging  reduces discrimination both in variance and discrimination in noise.
\end{proof}

\section{Experiment}
In this section,  we demonstrate our framework by performing experiments on real-world and synthetic datasets.
The goal of our experiments is three folds. The first is to show how the framework makes use of unlabeled data to achieve a better trade-off between accuracy and discrimination.
The second is to explore the impact of factors, such as ensemble times and sampling size, on the training results. 
And, third, we show the distinct difference in discrimination level when the model is tested with discrimination test dataset and fair test dataset.


\subsection{Experiments on Real Data}
 The aim of real-world datasets is to assess the effectiveness of our method to achieve a better trade-off between accuracy and discrimination with unlabeled data. We also show the benefit of ensemble learning, the impact of the sampling size, and the comparison with other methods.
 
\subsubsection{Experimental Setup}
\paragraph{Dataset}
The experiments involve three real-world datasets: the Health dataset \footnote{https://foreverdata.org/1015/index.html}, the Bank dataset \footnote{https://archive.ics.uci.edu/ml/datasets/bank+marketing}, the Adult dataset \footnote{http://archive.ics.uci.edu/ml/datasets/Adult}.
\begin{itemize}
	\item The target of Health dataset is to predict whether people will spend any day in the hospital. 
	In order to convert the problem into the binary classification task, we simply predict whether people will spend any day in the hospital or not. 
	Here, 'Age' is the protected attribute and two groups are divided at $ \geq $65 years.
	After data pre-processing, the dataset contains 10000 records with 132 features.
	\item The Bank dataset contains a total of 31,208 records with 20 attributes and a binary label, which indicates whether the client has subscribed to a term deposit or not. Again,  'Age' is the protected attribute.
	\item The target of Adult dataset is to predict whether people's income is larger than 50K dollars or not, and we consider "Gender" as the protected attribute. After data pre-processing, the dataset contains  48,842 records with 18 features.
	
\end{itemize}
\paragraph{Parameters}
The protected attribute is excluded from the prediction model during the training to ensure equity across groups. The protected attribute is only used to evaluate the discrimination measurement in the testing phrase. In the above of three real-world datasets, data are all labeled. First, we split the whole dataset randomly into two halves: one half is used as labeled dataset, and we remove the labels from the other half to served as the unlabeled dataset. In the labeled data, we set the split rate $ s=0.8 $, which means $ 80\% $ of the data are used for training and $ 20\% $ of the data are used for testing. The sample size $ n_s $ equals the minimum size of four groups in three datasets.

The final result is an average of 50 results run in the new training datasets.
For each run,
we generate $ K=200 $ fair datasets and construct with $ K=200 $ base models to make the final predictions. 
We use 5-fold cross-validation on the original training dataset and test dataset. 
\paragraph{Baseline}
Given our method is a pre-processing method, we compare it to two other pre-processing methods and the method without any fairness process.
\begin{itemize}
	\item Original (ORI): The original dataset is used for training without fairness guarantees. 
	\item Uniform Sampling (US) \cite{kamiran2012data} : The number of data points in each groups is equalized  through oversampling and/ undersampling. 
	\item Preferential Sampling (PS) \cite{kamiran2012data} : The number of data points in each groups is equalized by taking samples near the borderline data points. 
\end{itemize}

\subsubsection{Trade-off Between Accuracy and Discrimination}

\begin{figure}[ht]
	\begin{minipage}[b]{0.49\linewidth}
		\centering	
		\includegraphics[scale=0.22]{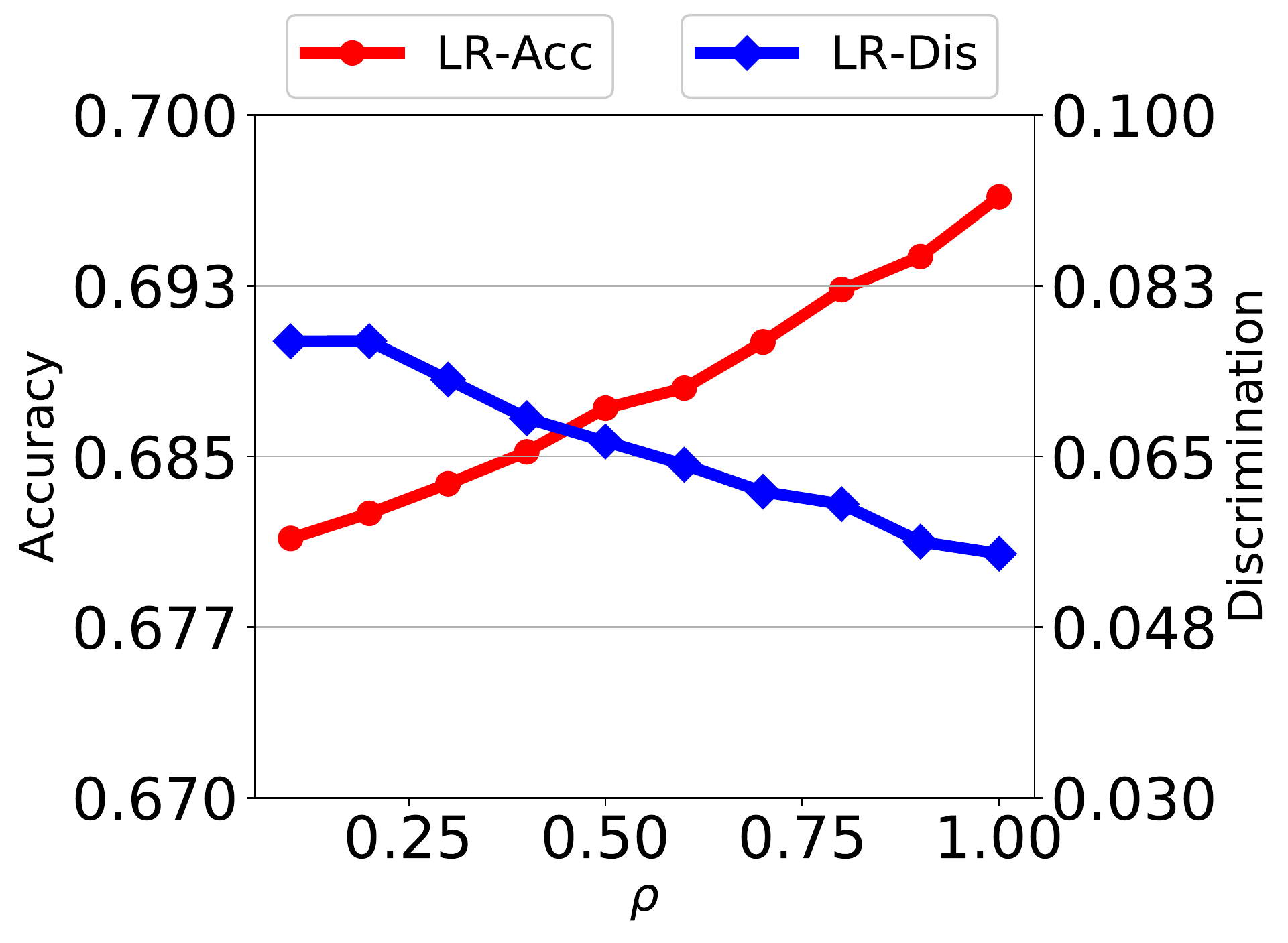}
		\centerline{(a) LR-Health}
	\end{minipage}
	\begin{minipage}[b]{0.49\linewidth}
		\centering
		\includegraphics[scale=0.22]{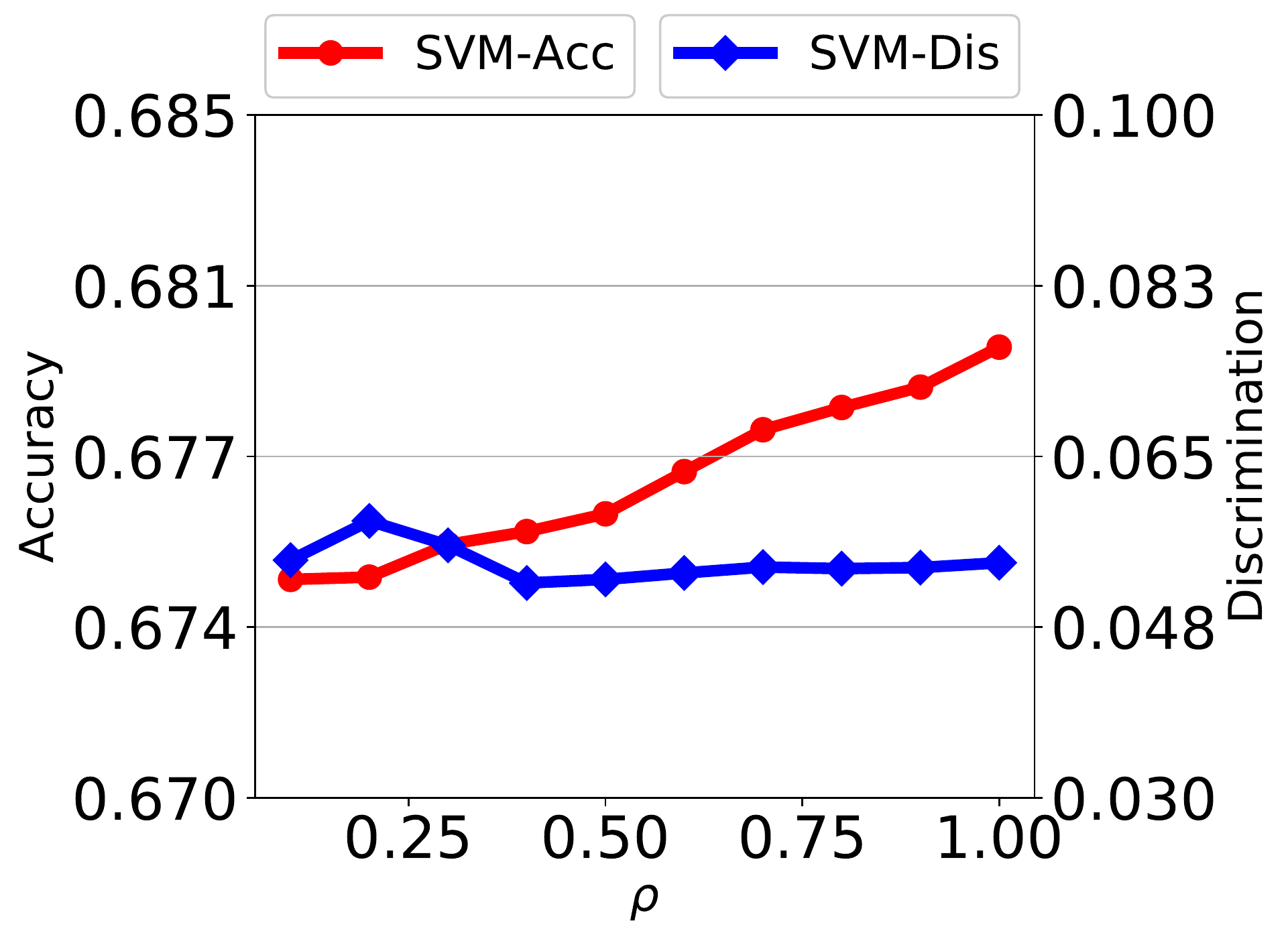}
		\centerline{(b) SVM-Health}
	\end{minipage}
	
	\begin{minipage}[b]{0.49\linewidth}
		\centering	
		\includegraphics[scale=0.22]{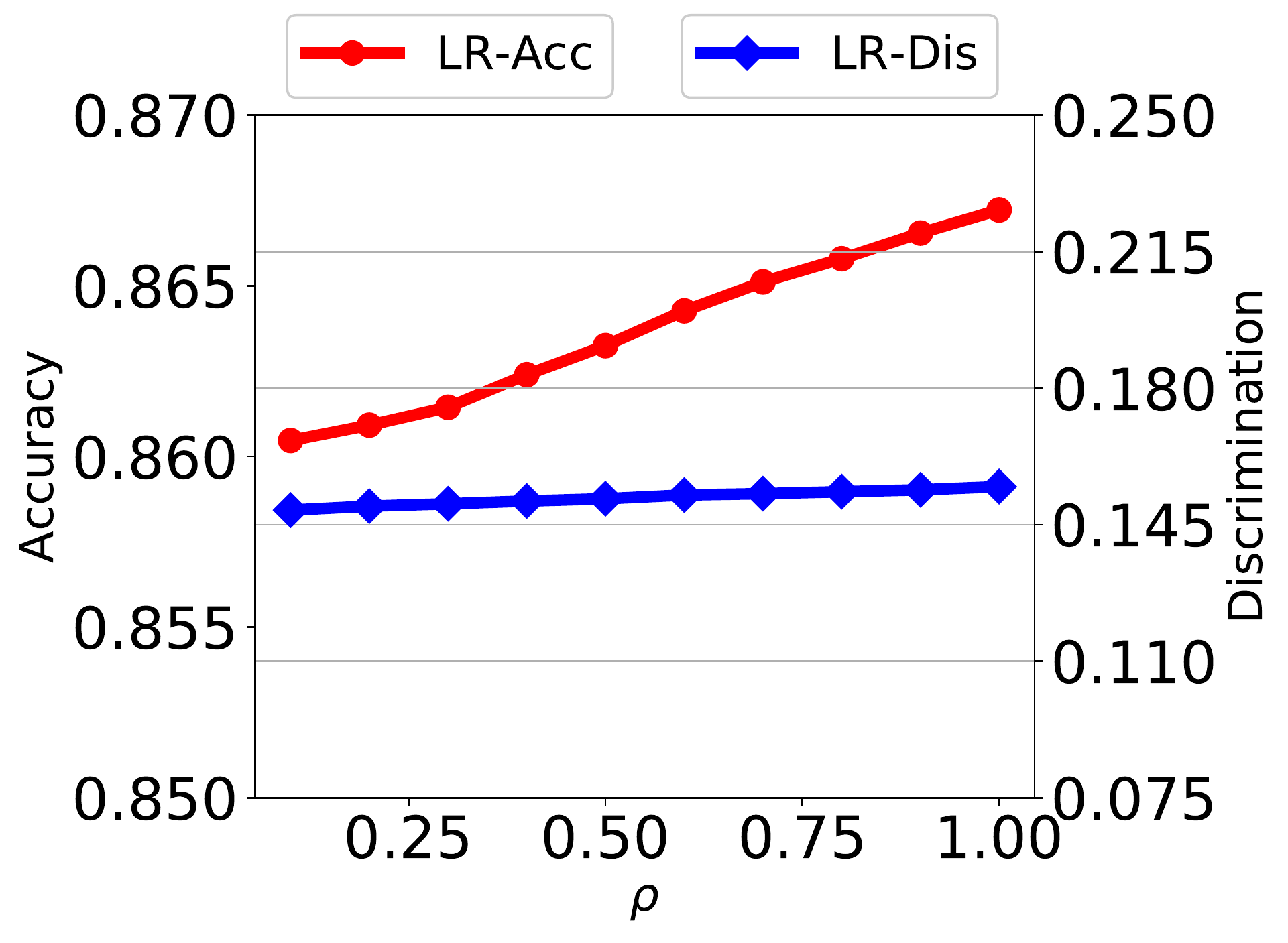}
		\centerline{(c) LR-Bank}
	\end{minipage}
	\begin{minipage}[b]{0.49\linewidth}
		\centering
		\includegraphics[scale=0.22]{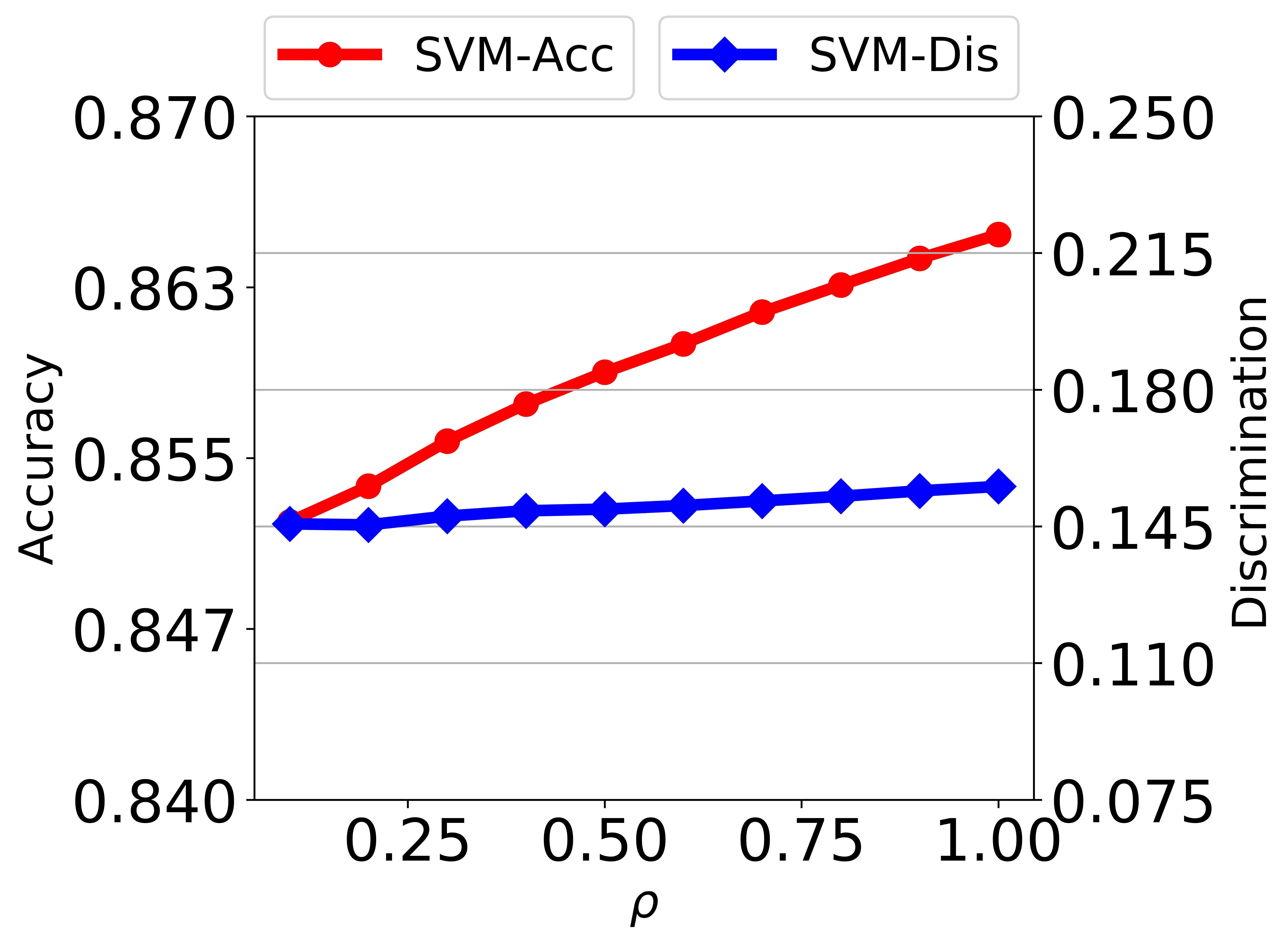}
		\centerline{(d) SVM-Bank}
	\end{minipage}

	\begin{minipage}[c]{0.49\linewidth}
	\centering	
	\includegraphics[scale=0.22]{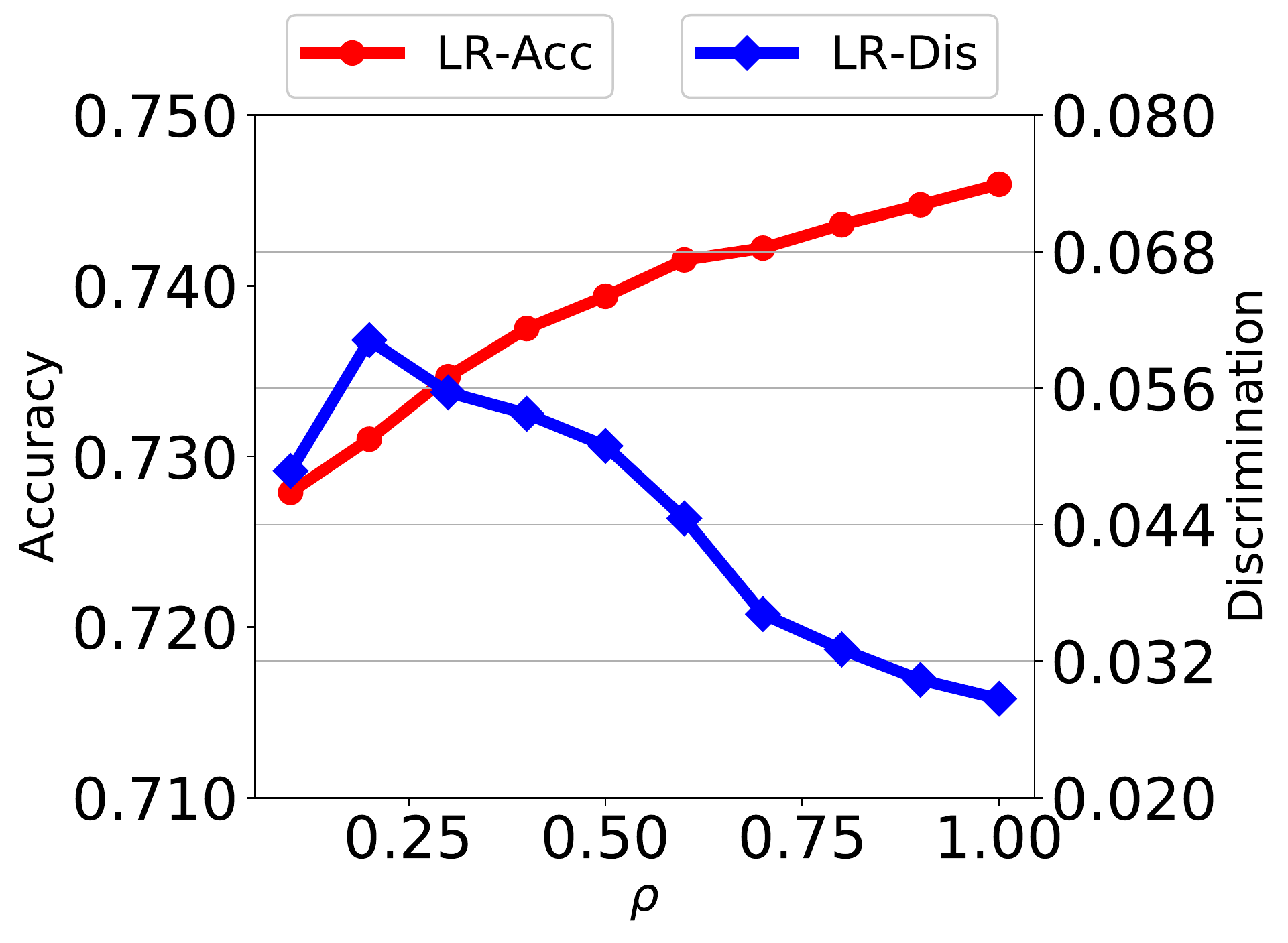}
	\centerline{(e) LR-Adult}
	\end{minipage}
	\begin{minipage}[e]{0.49\linewidth}
	\centering
	\includegraphics[scale=0.22]{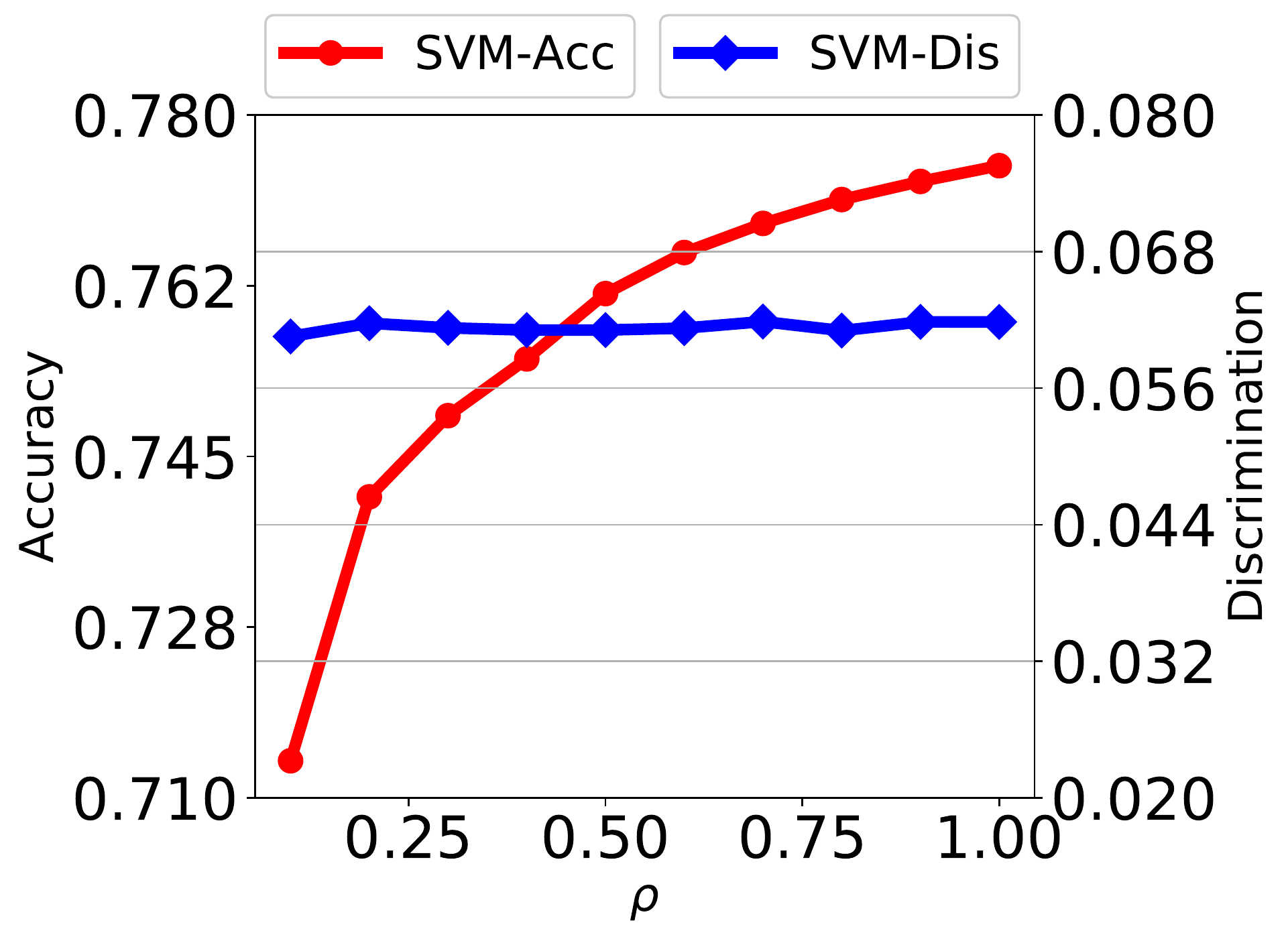}
	\centerline{(f) SVM-Adult}
	\end{minipage}
	\caption{The trade-off between accuracy (Red) and discrimination level (Blue).  (a) LR in Health dataset; (b) SVM in Health dataset; (c) LR in Bank dataset; (d) SVM in Bank dataset;  (e) LR in Adult dataset; (f) SVM in Adult dataset. The X-axis is the sample ratio $ \rho $, which denotes that the percentage of $ \rho $ unlabeled data are sampled from the unlabeled dataset and then pseudo labeled for training.  }	
\end{figure}
Figure 2 shows the accuracy and discrimination level varies given different sample ratio $ \rho $ with logistic regression (LR) and support vector machine (SVM)  on three datasets. 
As shown, accuracy generally increases with a growing size of unlabeled data. For example,  LR has an accuracy of around 0.728  when $ \rho=0.1 $ with the Adult dataset, which increases to 0.745  when $ \rho=1 $. 
This indicates that the unlabeled data helps to improve the accuracy to some extent. Also, we note that accuracy relates to the training models and the choice of training models relates to the datasets. 
The discrimination level has different performances in different training models.
For example, with the Adult dataset, the discrimination level initially increases and then steadily decreases till the end in LR. 
The discrimination level is steady and has a slight increase in SVM. 
This observation indicates that unlabeled data can help to reduce the discrimination for some models, like LR. Similar to accuracy, the discrimination level  
relates to the training models and our experiments show that LR is more friendly in discrimination than SVM. The choice of sample ratio depends on the quality of the dataset itself as well as the requirement of the learning task. Accuracy could be improved with unlabeled data, while discrimination level depends on the reduction of discrimination in variance and increase of discrimination in noise that unlabeled data could bring in the training. 

\begin{figure}[ht]
		\begin{minipage}[b]{0.49\linewidth}
		\centering	
		\includegraphics[scale=0.22]{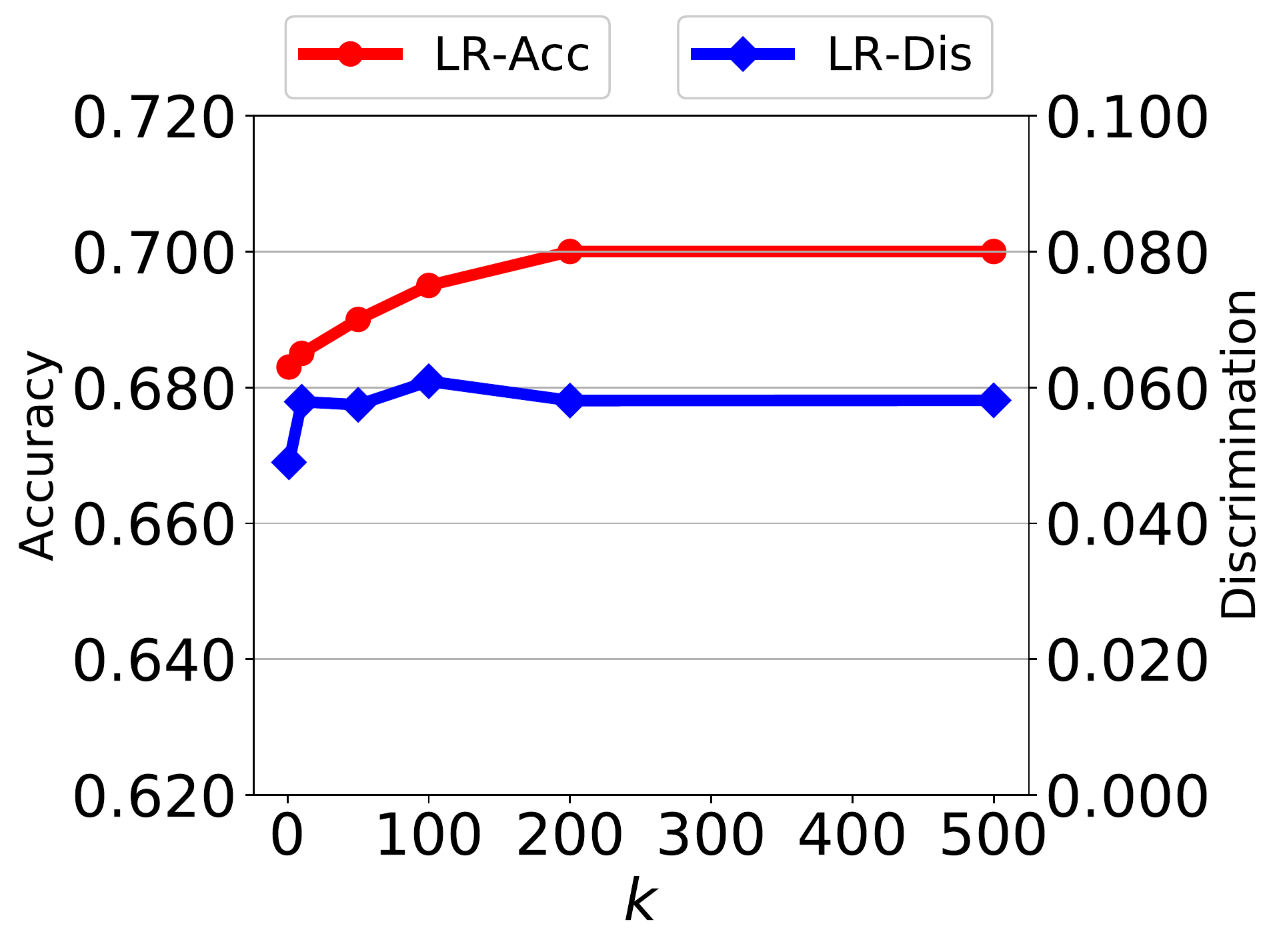}
		\centerline{(a) LR-Health}
	\end{minipage}
	\begin{minipage}[b]{0.49\linewidth}
		\includegraphics[scale=0.22]{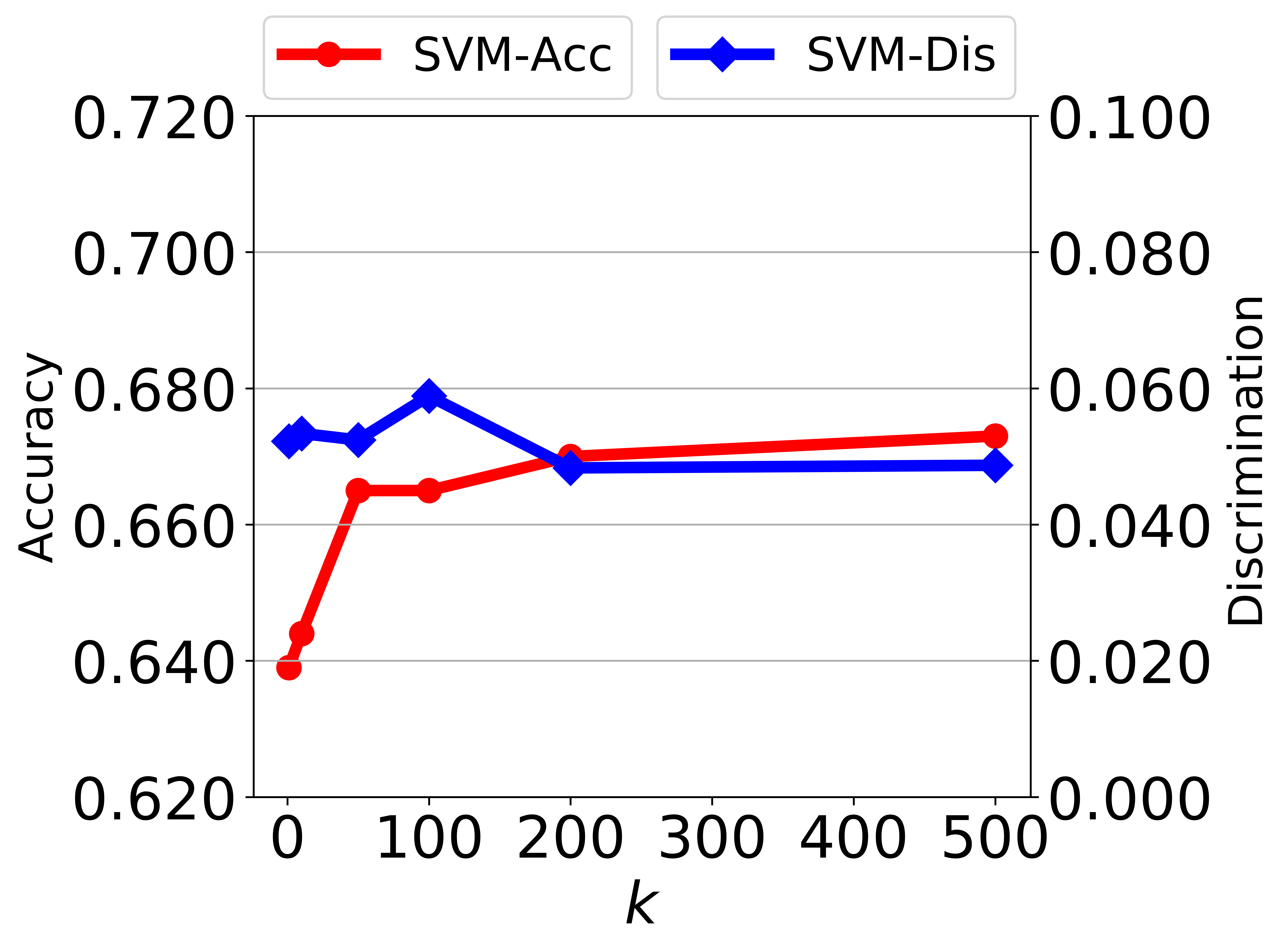}
		\centerline{(b) SVM-Health}
	\end{minipage}

	\begin{minipage}[b]{0.49\linewidth}
	\centering	
	\includegraphics[scale=0.22]{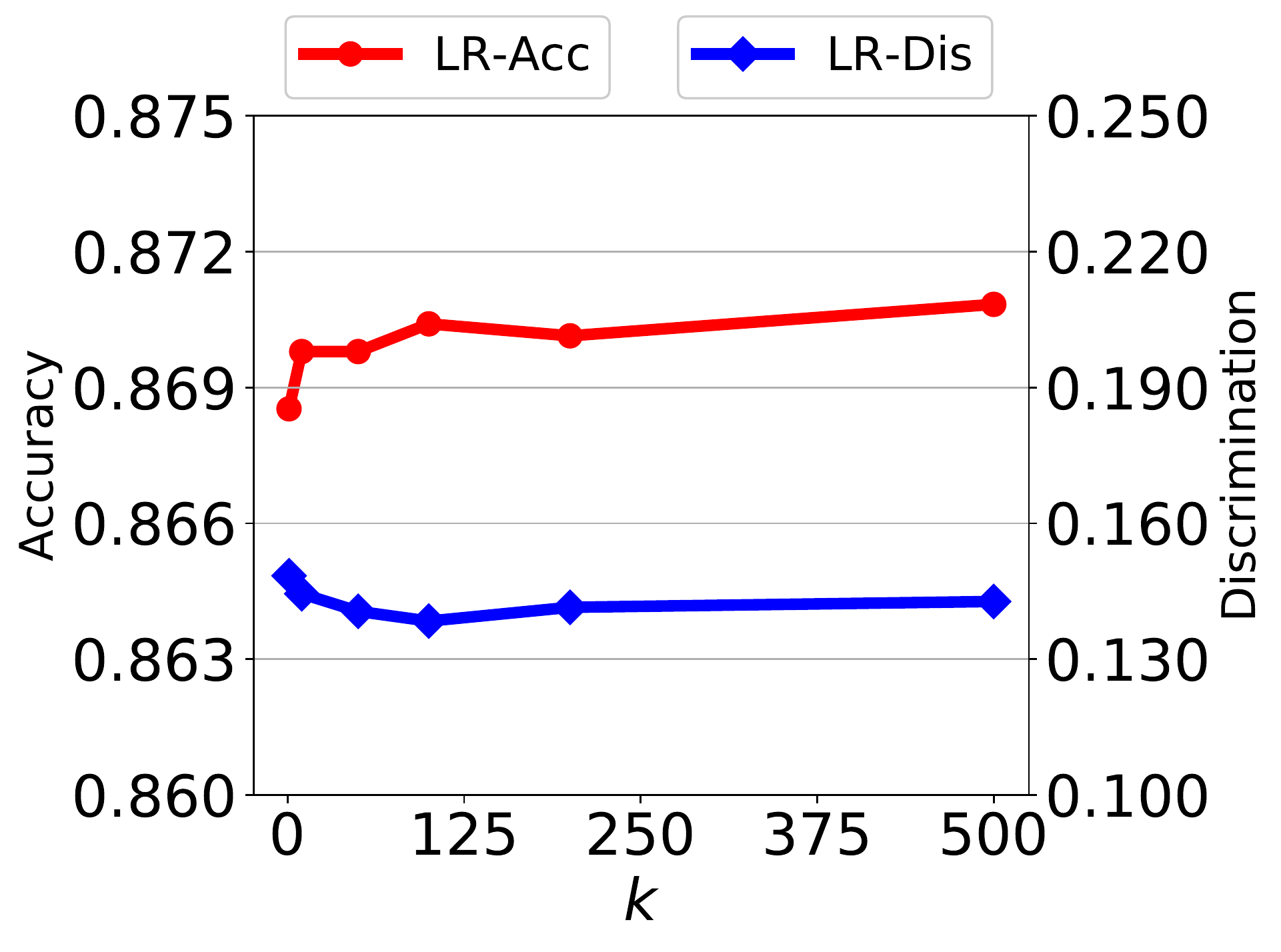}
	\centerline{(c) LR-Bank}
	\end{minipage}
	\begin{minipage}[b]{0.49\linewidth}
	\includegraphics[scale=0.22]{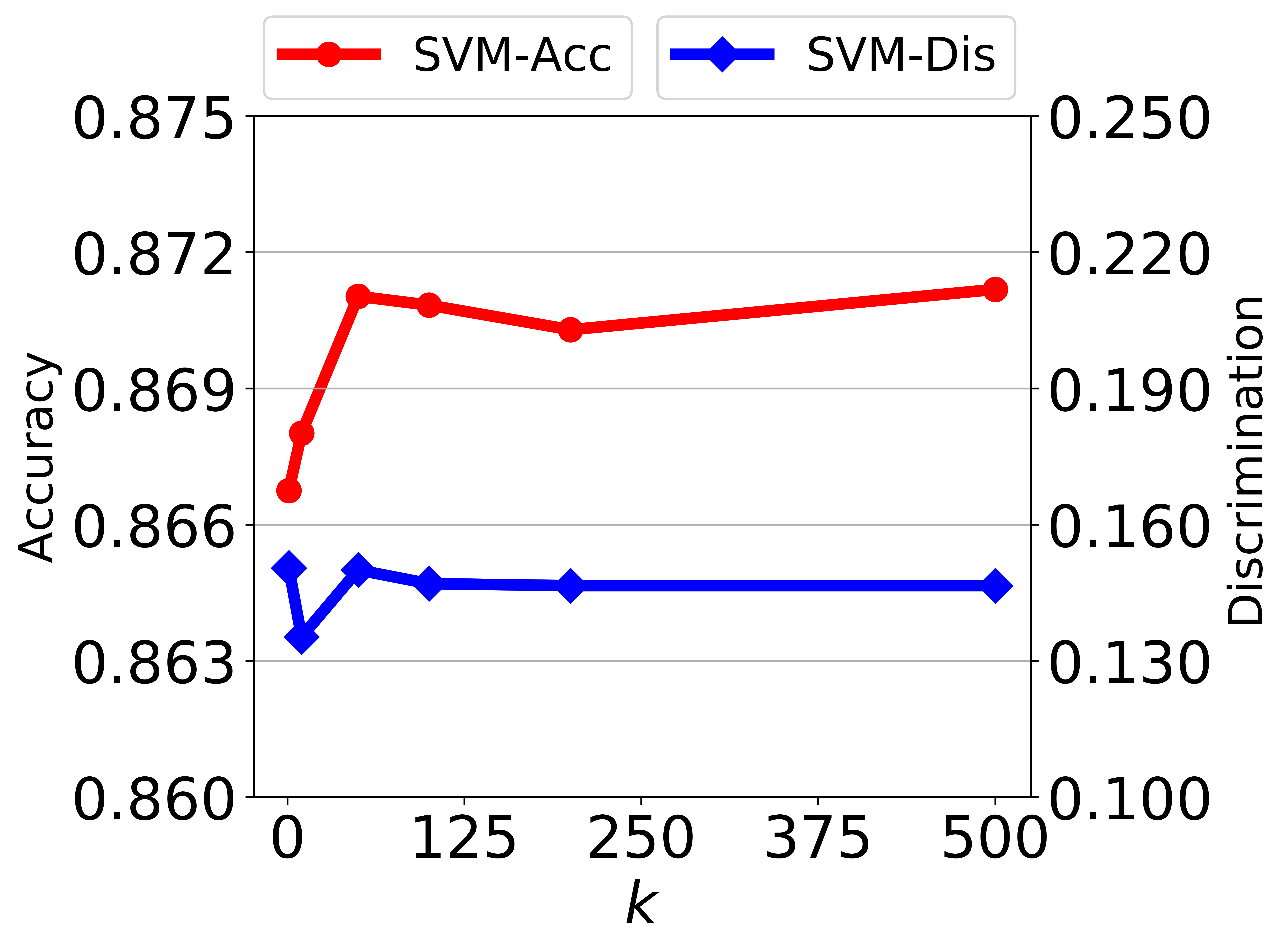}
	\centerline{(d) SVM-Bank}
	\end{minipage}

	\begin{minipage}[b]{0.49\linewidth}
		\centering	
		\includegraphics[scale=0.22]{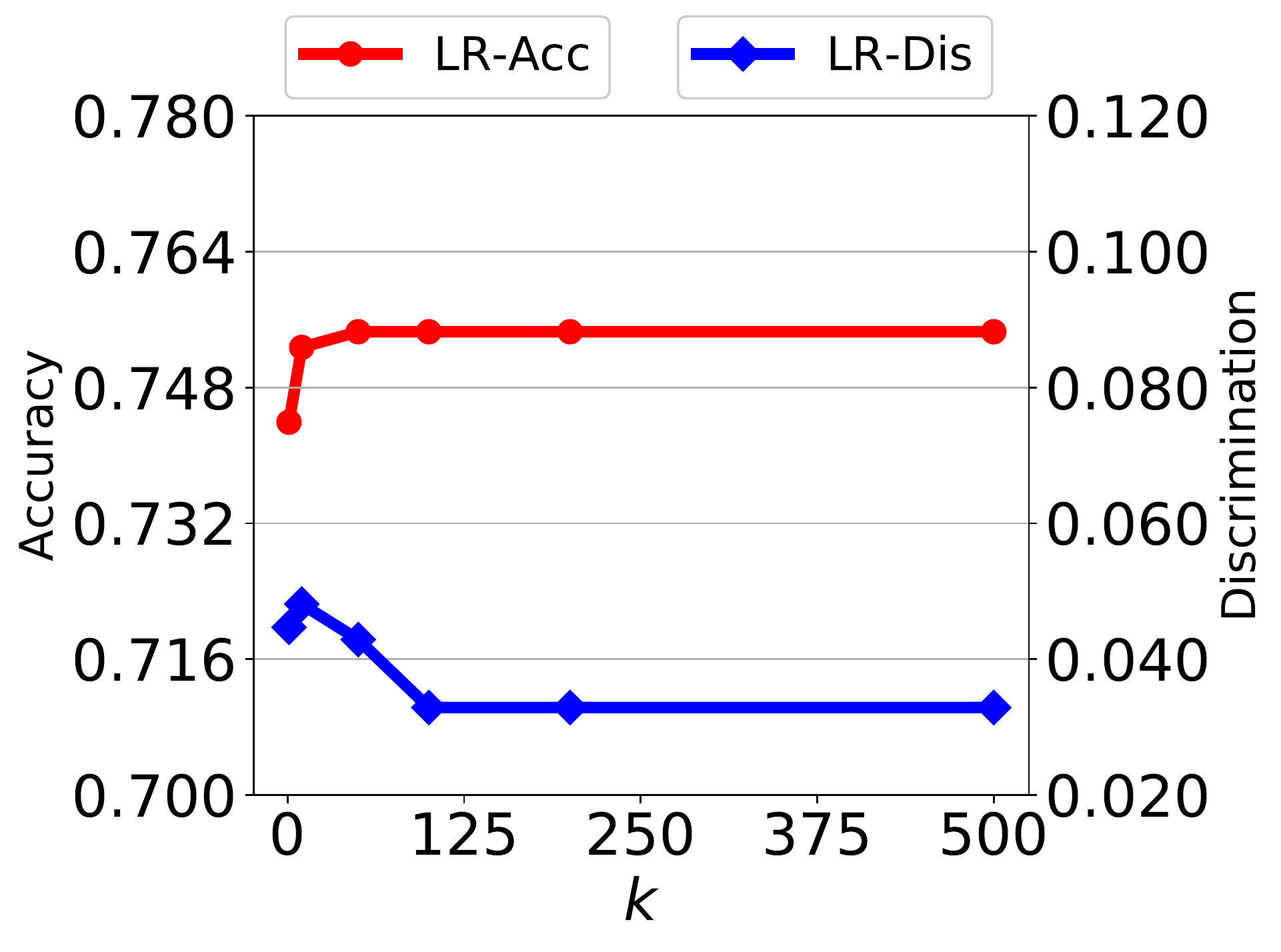}
		\centerline{(e) LR-Adult}
	\end{minipage}
	\begin{minipage}[b]{0.49\linewidth}
		\includegraphics[scale=0.22]{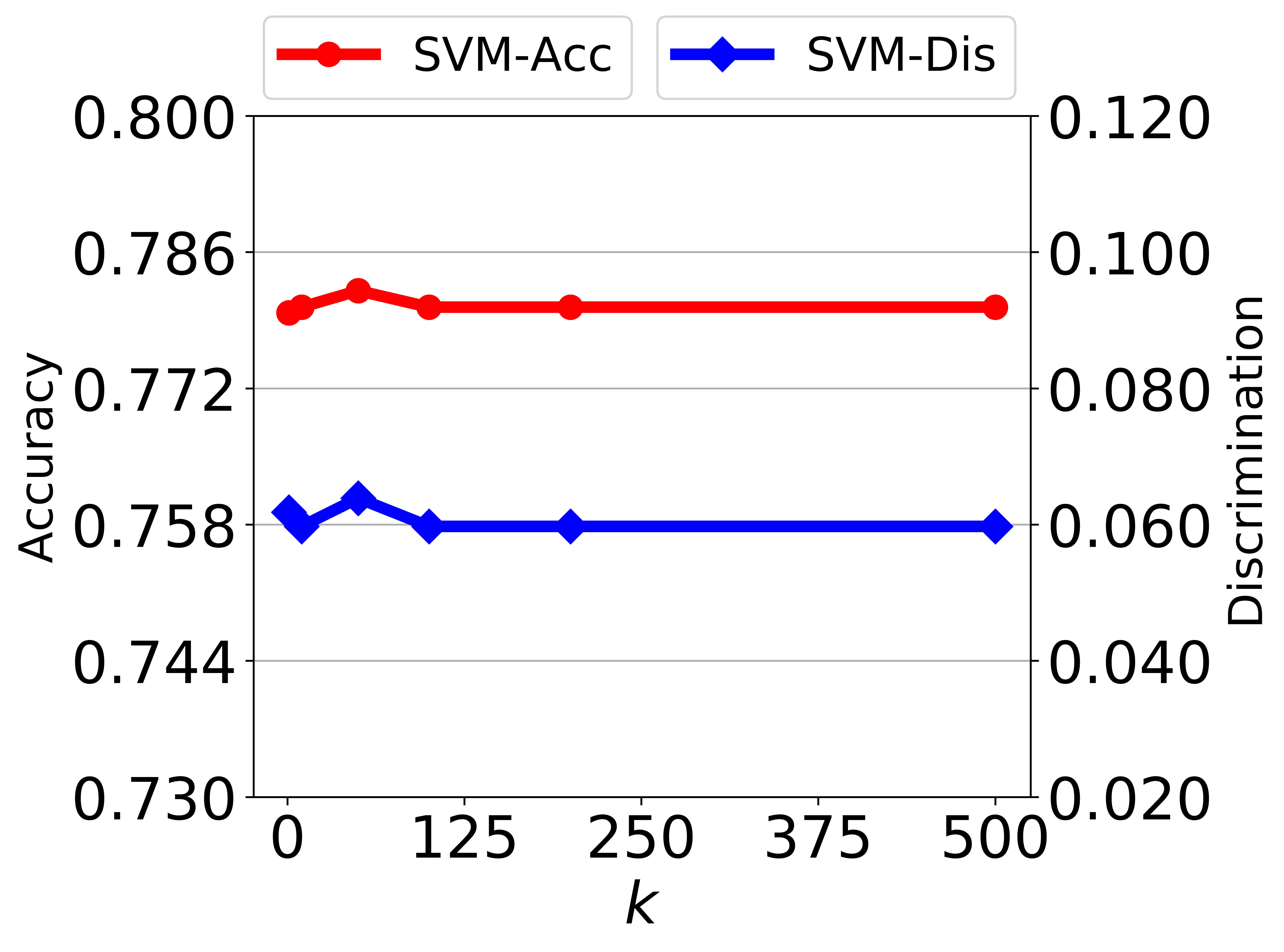}
		\centerline{(f) SVM-Adult}
	\end{minipage}

	\caption{ The impact of ensemble learning on the accuracy (Red) and discrimination level (Blue) on (a) LR in Health dataset; (b) SVM in Health dataset; (c) LR in Bank dataset; (d) SVM in Bank dataset;  (e) LR in Adult dataset; (f) SVM in Adult dataset. Initially, there is not obvious link between accuracy and discrimination level. However, as the ensemble size grows, the accuracy and discrimination level begin to converge. Each point is an average of 50 times.}
\end{figure}

\subsubsection{The Impact of Ensemble Learning}
Figure 3 shows the impact of ensemble learning on accuracy and discrimination level with LR and SVM on three datasets. 
In ensemble learning, we sample percentage of $ \rho=1 $  unlabeled data from the unlabeled dataset, and generate the new training dataset. 
With LR, the accuracy typically increases then steadies till the end, whereas, with SVM accuracy fluctuates before steadying at some lower, equal or higher rate. This is because the errors in variance and noise reduce as the ensemble size increases. 

In terms of discrimination levels, both methods show fluctuations at first before stabilizing on all three datasets.
The changes in discrimination levels have no obvious correlations to accuracy prior to convergence. This is reasonable because training results having the same accuracy does not mean the same discrimination level. Also, without a sufficient ensemble size, training on fair datasets will introduce some variance and noise to the final result. Overall, an ample ensemble size helps to improve accuracy and decrease discrimination. The appropriate ensemble size is $ K=200 $ or so. This is because accuracy increases and discrimination fluctuates before $ K=200 $, and broadly accuracy and discrimination become steady after $ K=200 $ for three datasets.

\subsubsection{The Impact of Sample Size}
\begin{figure}[ht]
	\begin{minipage}[b]{0.49\linewidth}
		\centering	
		\includegraphics[scale=0.22]{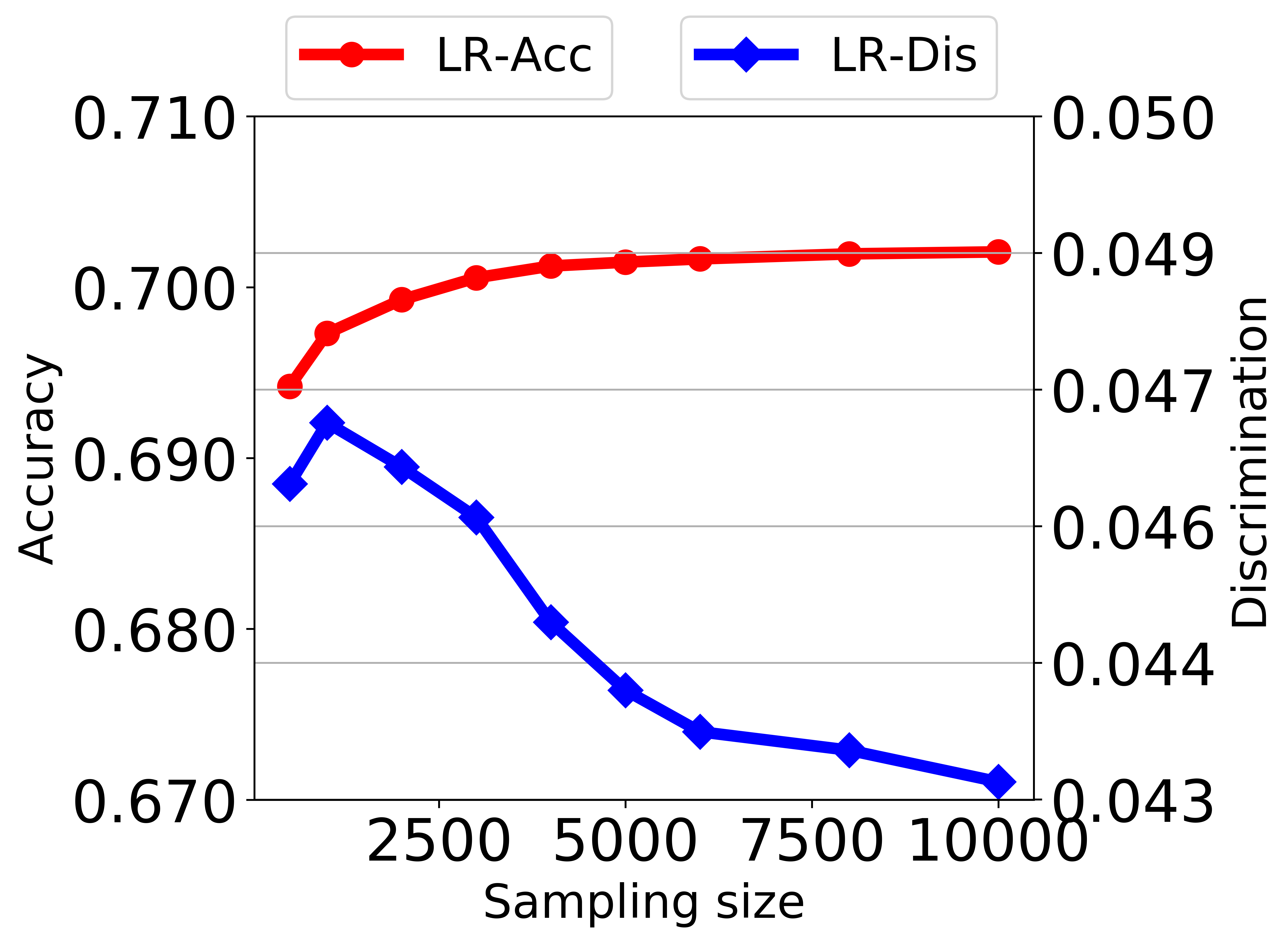}
		\centerline{(a) LR-Health}
	\end{minipage}
	\begin{minipage}[b]{0.49\linewidth}
		\centering
		\includegraphics[scale=0.22]{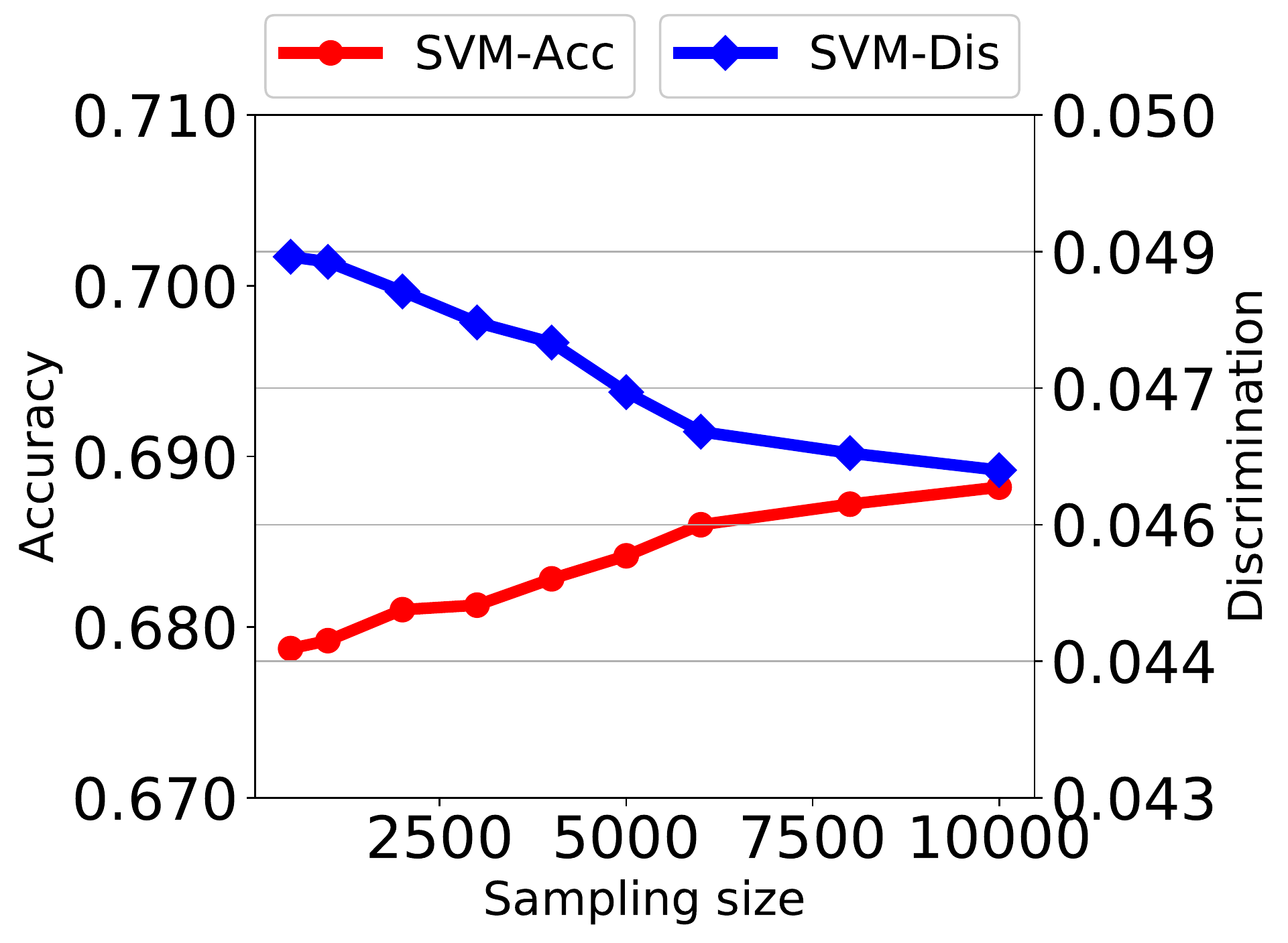}
		\centerline{(b) SVM-Health}
	\end{minipage}	

	\begin{minipage}[b]{0.49\linewidth}
	\centering	
	\includegraphics[scale=0.22]{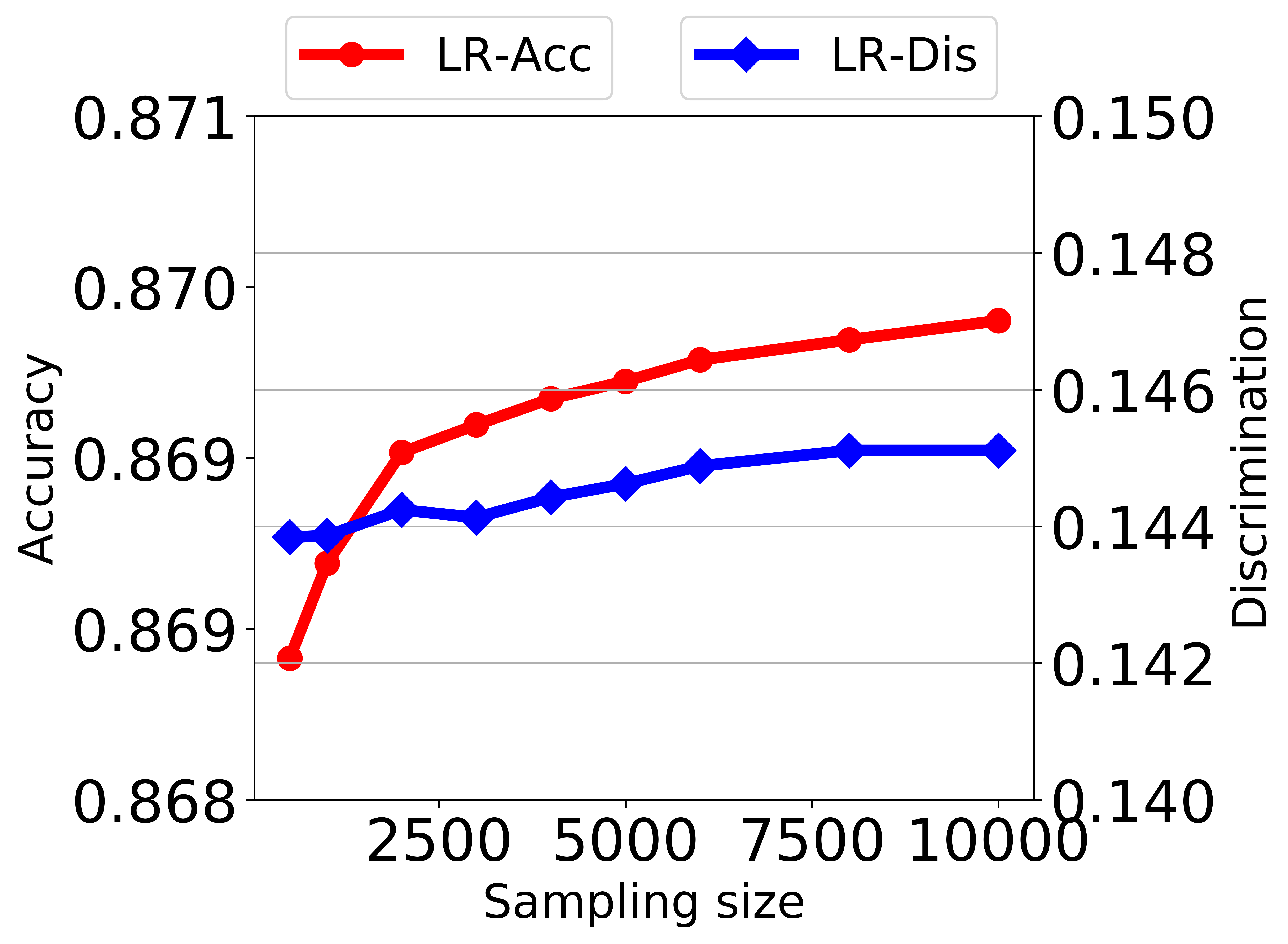}
	\centerline{(c) LR-Bank}
 	\end{minipage}
	\begin{minipage}[b]{0.49\linewidth}
	\centering
	\includegraphics[scale=0.22]{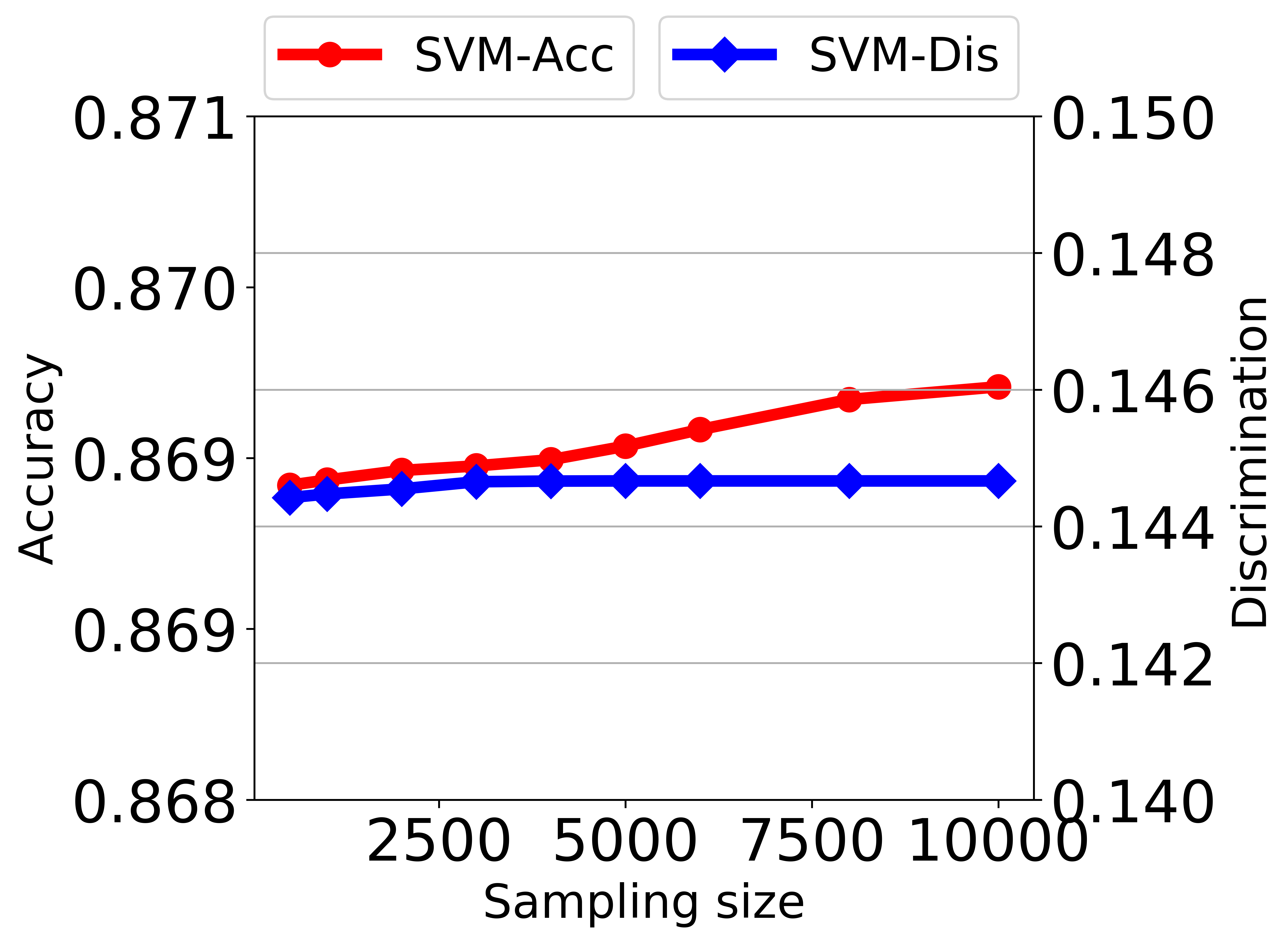}
	\centerline{(d) SVM-Bank}
    \end{minipage}

	\begin{minipage}[b]{0.49\linewidth}
		\centering	
		\includegraphics[scale=0.22]{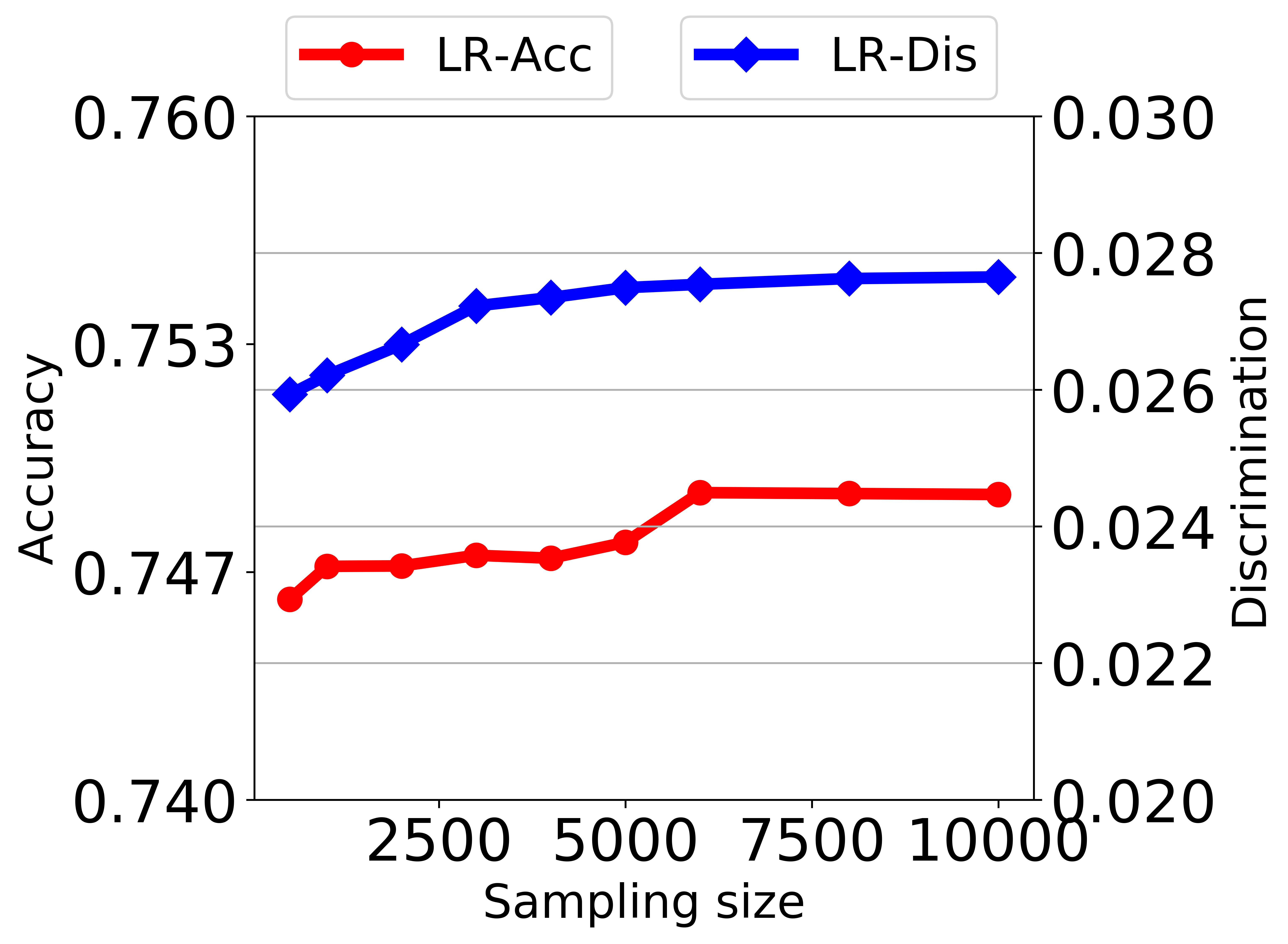}
		\centerline{(e) LR-Adult}
	\end{minipage}
	\begin{minipage}[b]{0.49\linewidth}
		\centering
		\includegraphics[scale=0.22]{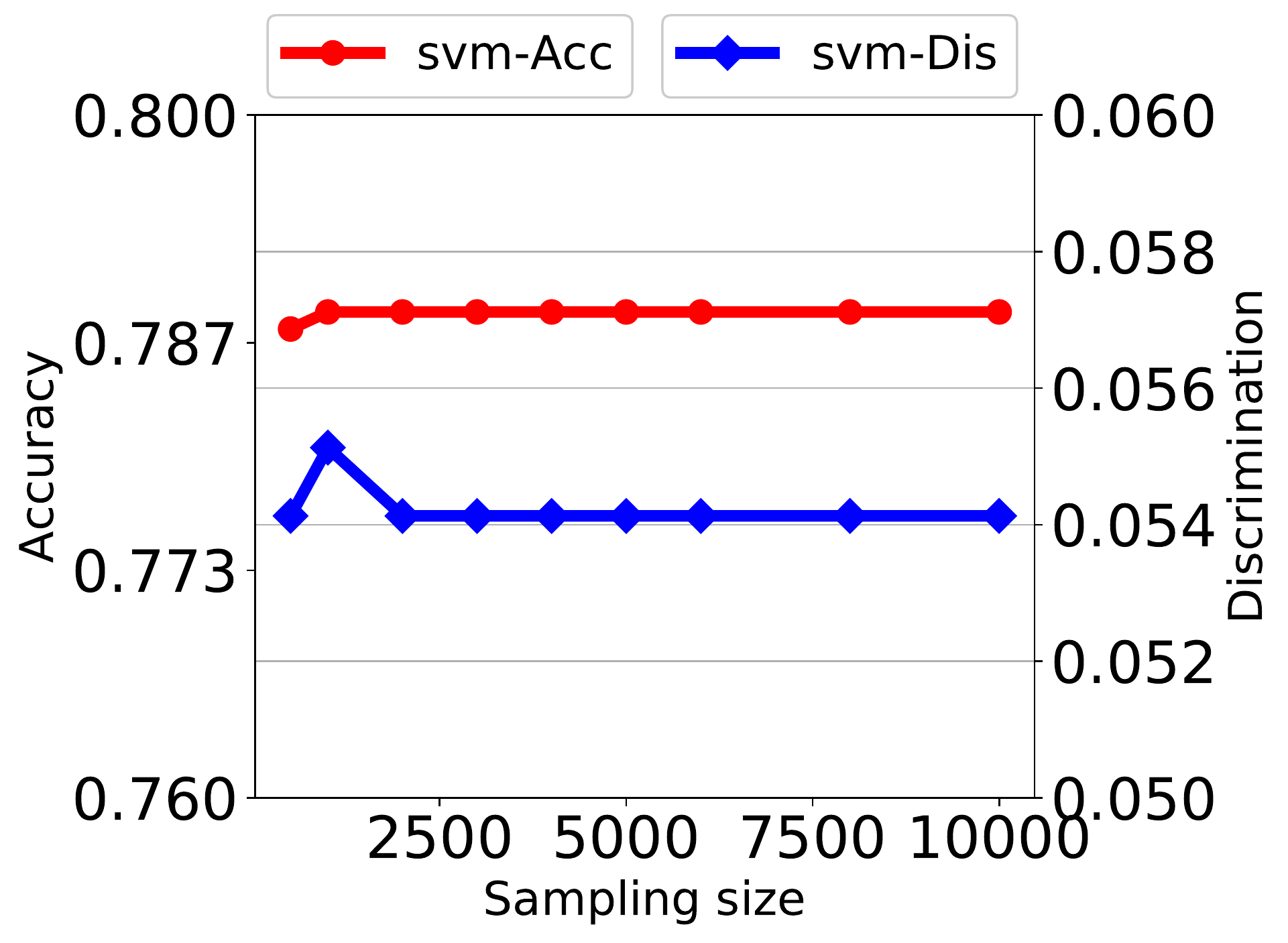}
		\centerline{(f) SVM-Adult}
	\end{minipage}	
	\caption{The impact of sample size on accuracy (Red) and discrimination level (Blue) on (a) LR in Health dataset; (b) SVM in Health dataset; (c) LR in Bank dataset; (d) SVM in Bank dataset; (e) LR in Adult dataset; (f) SVM in Adult dataset. An increasing in the sampling size leads to an increase in accuracy and may help to reduce discrimination level. }
\end{figure}
Figure 4 shows the impact of sample size on accuracy and discrimination level with LR and SVM on three datasets. Overall, it is observed that   accuracy increases quickly in the early stages and then becomes stable as the sample size grows. This is because more data help to improve the generalization ability, but extra data do not help when the amount of data is enough to fit the model.
Unlike accuracy, discrimination level depends on the amount of label noise that unlabeled data may bring when the sample size increases. For example, discrimination decreases in the Health dataset and increases a litter in the Bank dataset. 
This means that, with an increasing of sample size, little label noise is brought into the Health dataset, and consequently discrimination level decreases. Also, it is note that LR is more sensitive to sample size than SVM. The choice of sample size depends on the quality of the dataset and the training task requirement. Generally, a larger sample size can improve accuracy, reduce discrimination in bias and increase discrimination in noise.

\subsubsection{Comparison with other methods}
Figure 5 shows the results from a comparison of our proposed FS method with and the other three schemes in terms of the accuracy and discrimination level on the three datasets.
The training dataset of other methods is the original training dataset and the training dataset of our method is the new training dataset that consists of the original training dataset and pseudo labeled dataset ($\rho$=1). The test dataset is the same.
The results show that our method is able to push the discrimination to very low values while achieving a fairly high accuracy comparing with other schemes. Specifically, on the Adult dataset,  the discrimination level under LR is around 0.215 with the original method and around 0.022 with the preferential sampling method, and the proposed FS method can decrease discrimination to 0.019 with a better accuracy than the preferred sampling method. 
This indicates that the proposed FS method is able to reduce the discrimination better than other methods. 
\begin{figure}[ht]
	\begin{minipage}[b]{0.49\linewidth}
		\centering	
		\includegraphics[scale=0.22]{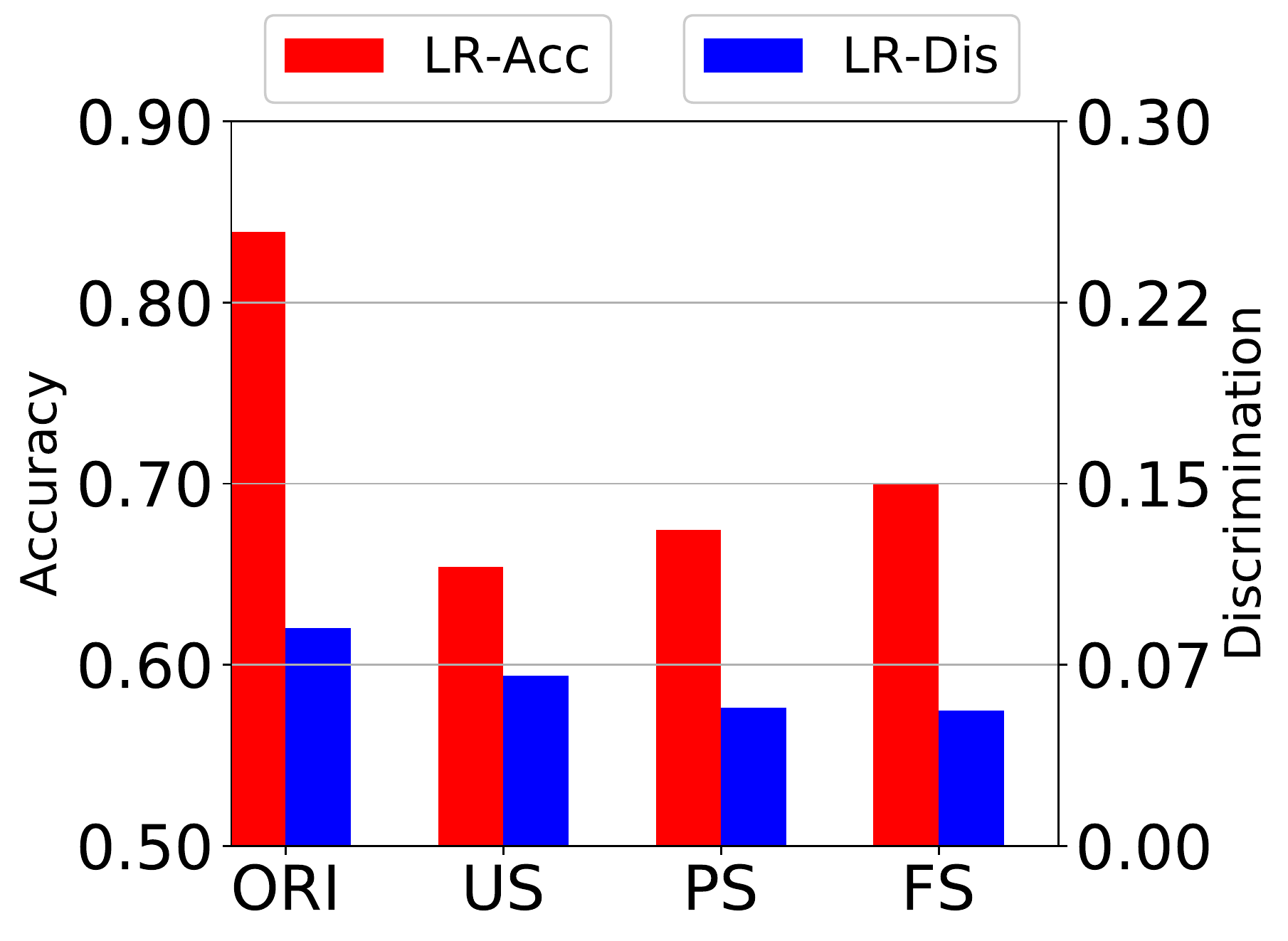}
		\centerline{(a) LR-Health}
	\end{minipage}
	\begin{minipage}[b]{0.49\linewidth}
		\centering
		\includegraphics[scale=0.22]{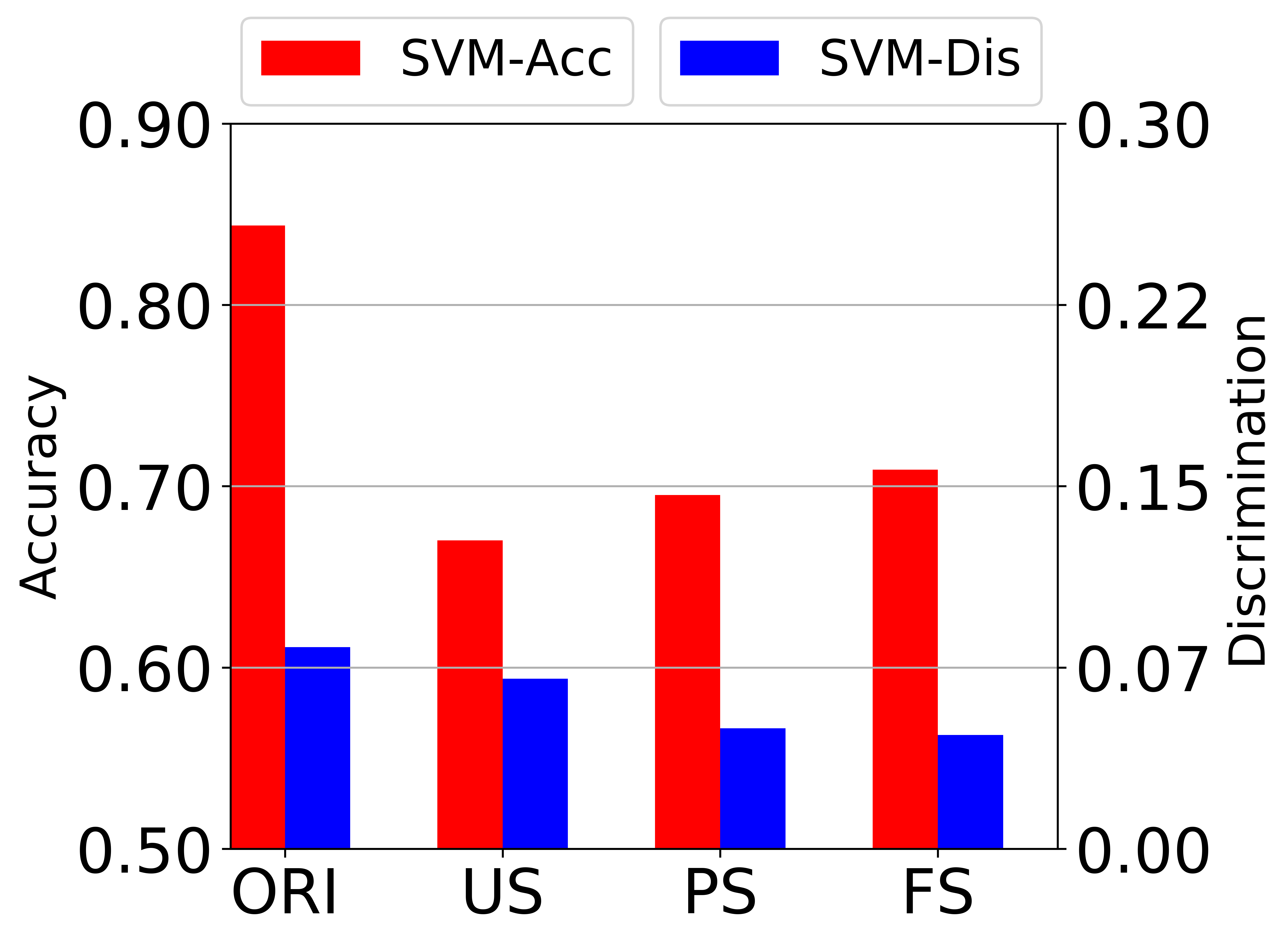}
		\centerline{(b) SVM-Health}
	\end{minipage}
	
	\begin{minipage}[b]{0.49\linewidth}
		\centering	
		\includegraphics[scale=0.22]{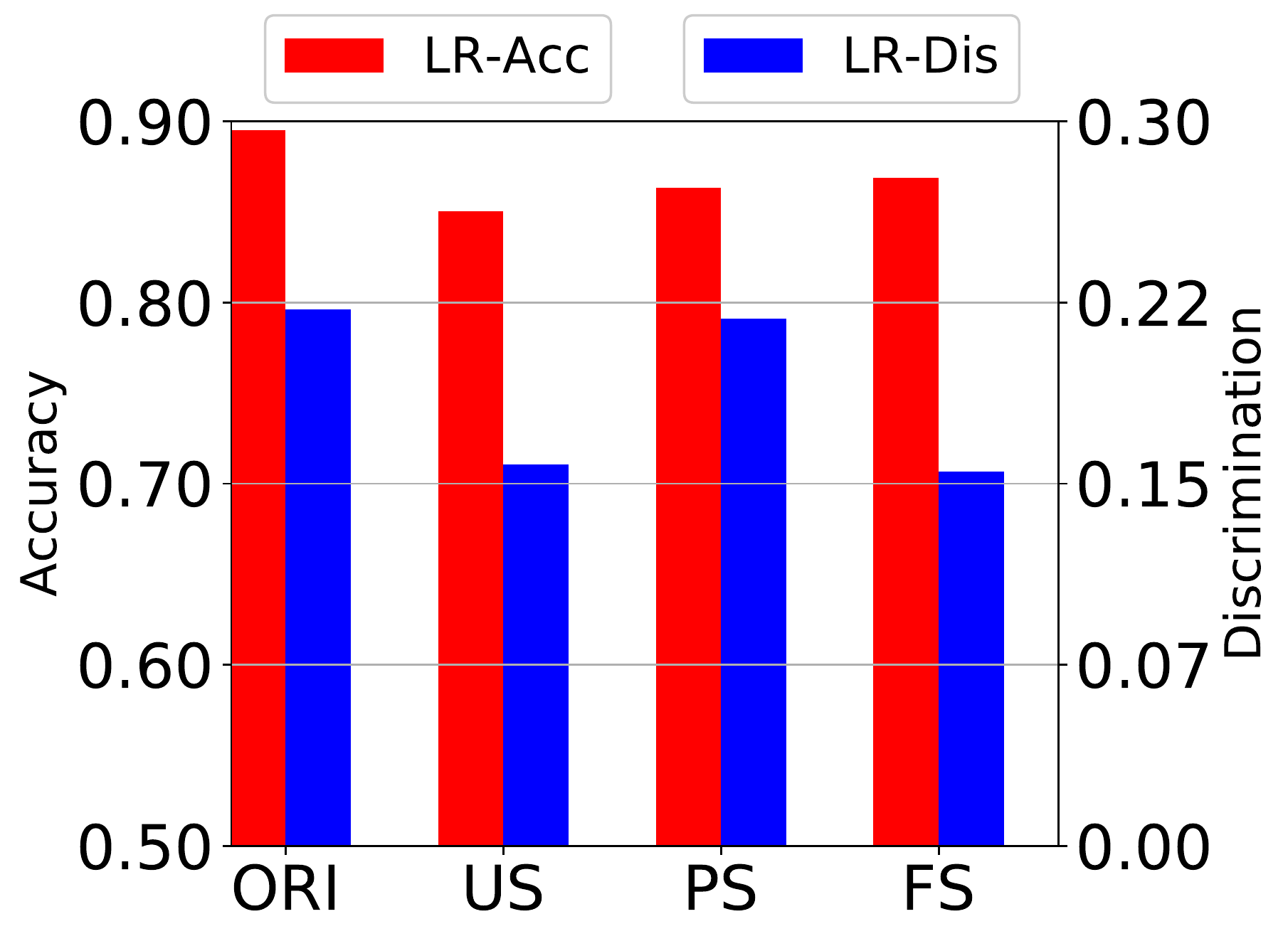}
		\centerline{(c) LR-Bank}
	\end{minipage}
	\begin{minipage}[b]{0.49\linewidth}
		\centering
		\includegraphics[scale=0.22]{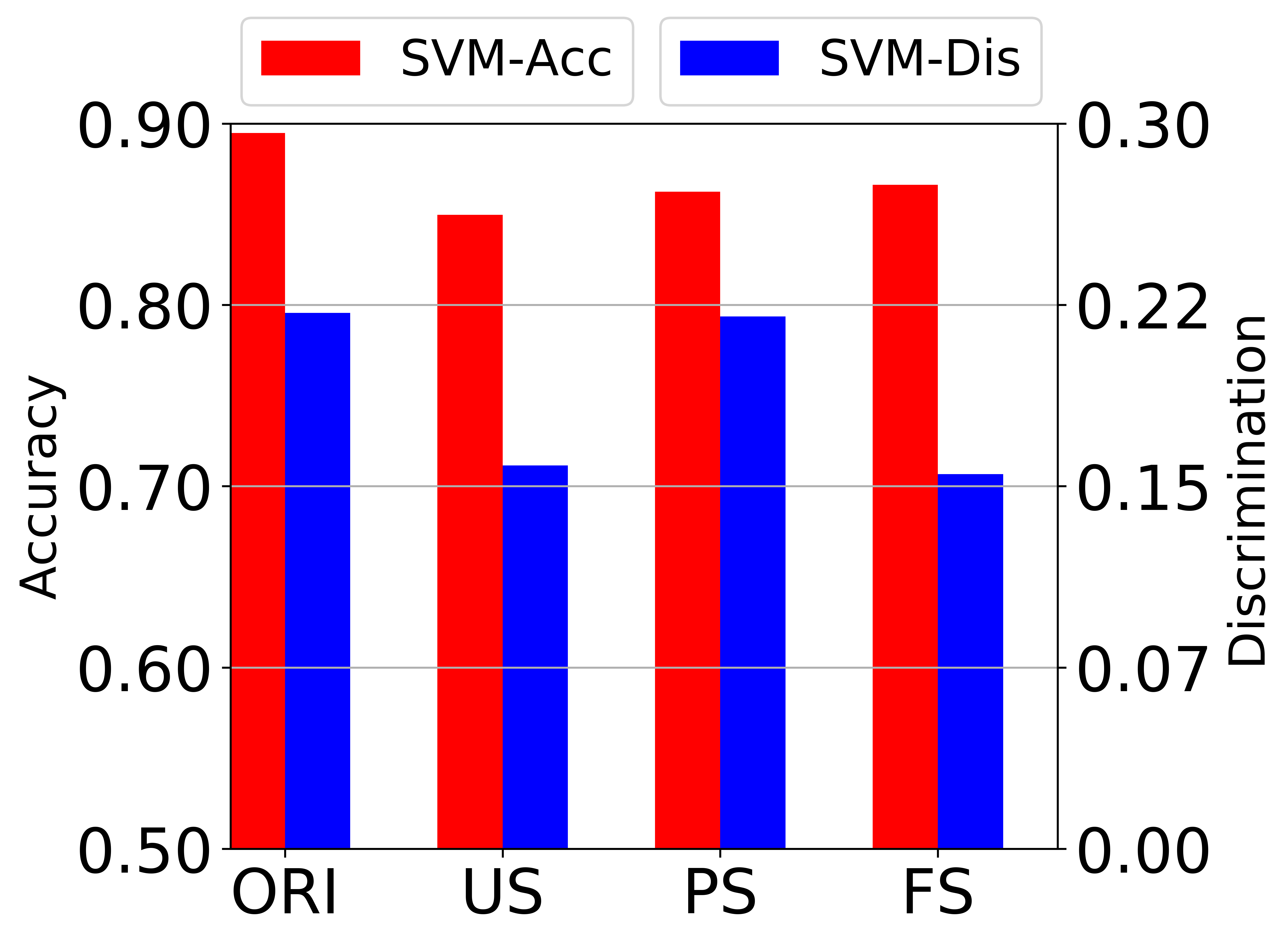}
		\centerline{(d) SVM-Bank}
	\end{minipage}
	
	\begin{minipage}[b]{0.49\linewidth}
		\centering	
		\includegraphics[scale=0.22]{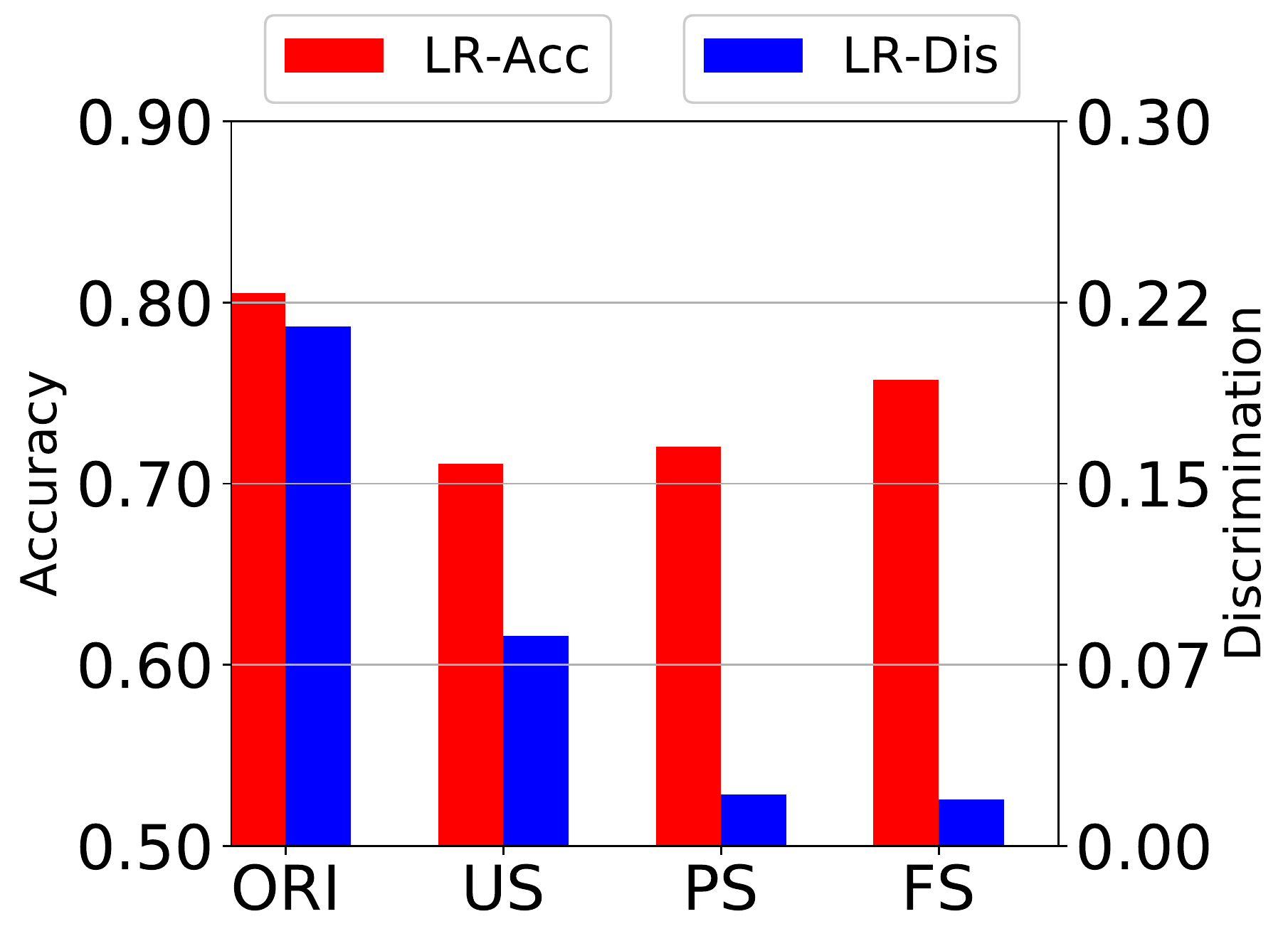}
		\centerline{(e) LR-Adult}
	\end{minipage}
	\begin{minipage}[b]{0.49\linewidth}
		\centering
		\includegraphics[scale=0.22]{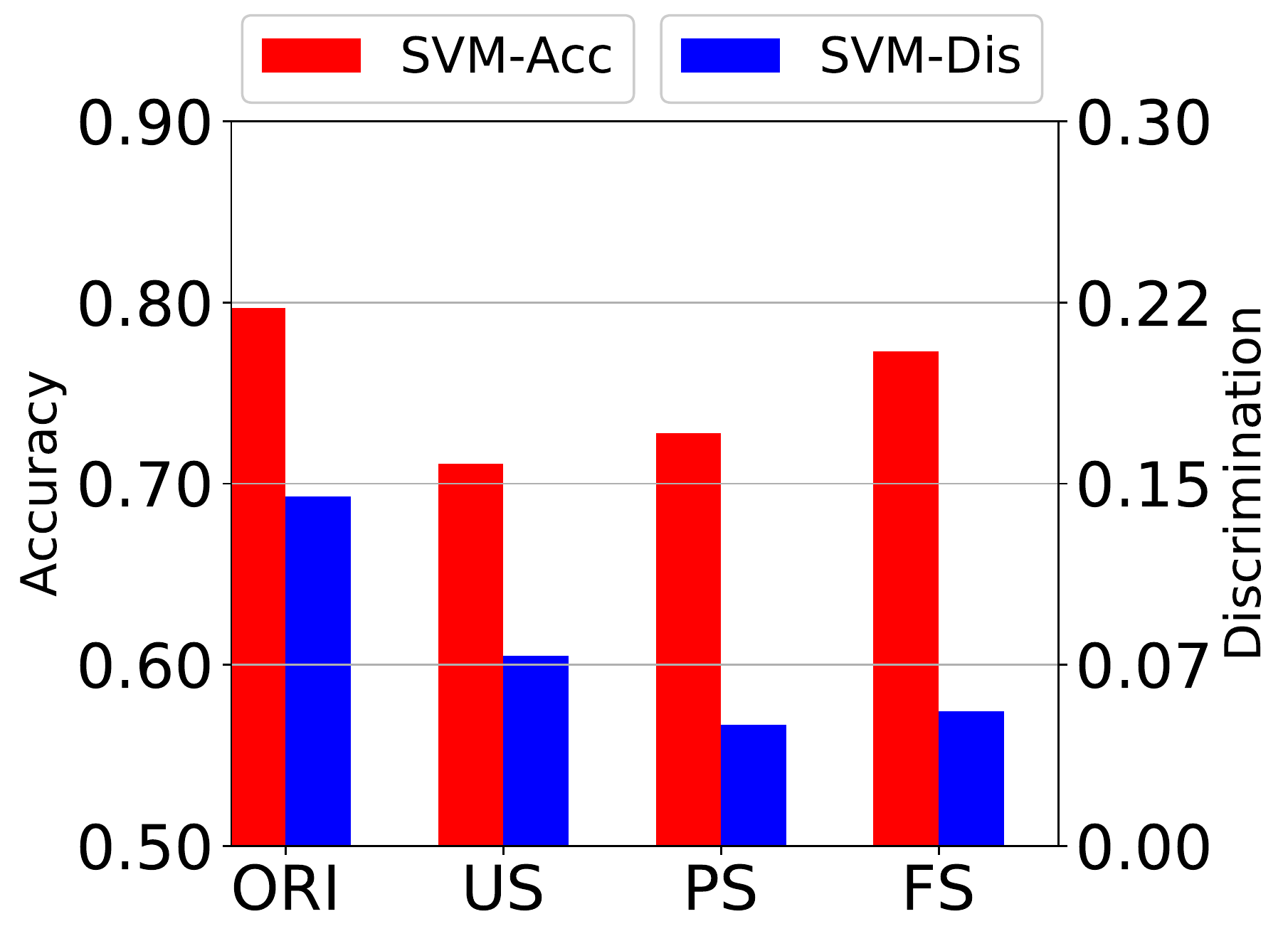}
		\centerline{(f) SVM-Adult}
	\end{minipage}	
	\caption{Comparison with original scheme (ORI), uniform sampling (US) and preferential sample (PS) with (a) LR in Health dataset; (b) SVM in Health dataset; (c) LR in Bank dataset; (d) SVM in Bank dataset; (e) LR in Adult dataset; (f) SVM in Adult dataset. With the fairness-enhanced sampling method (FS), discrimination decreases without much cost of accuracy or accuracy increases without much cost of discrimination. }
\end{figure}	
.

\subsection{Experiments on Synthetic Data}
We first describe how to generate synthetic datasets and the goal of synthetic datasets is to show the effectiveness of our method in the discriminatory test dataset and fair test dataset. Here, the discriminatory test dataset refers to the test dataset whose data points are not equally presented in each group, and the fair test dataset refers to the test dataset whose data points are equally presented in each group.
We show the distinct difference of discriminatory on two types of test datasets.
\subsubsection{Synthetic Data Setup}
We generate 22,000 binary class labels and a protected attribute $ a $ with a uniform random distribution, and assign a 2-dimensional feature vector to each label by drawing samples from two different Gaussian distributions: $ p(x|y=1)=N([2;2],[5,1;1,5]) $ and $ p(x|y=-1)=N([-2;-2],[10,1;1,3]) $. The size of each group in the synthetic dataset is roughly the same.
Then we randomly sample 2,000 data points from the synthetic dataset as a fair test dataset, and split the remaining dataset randomly into two halves: one half is to be used as the labeled dataset and the other half with labels removed to serve as the unlabeled dataset.

Note that the synthetic dataset is a fair dataset, and the discriminatory dataset is generated by calibrating data points in the group $G_{PP}$ based on the synthetic dataset.
Discriminatory dataset 1 (DA 1) is generated by sampling 2,000 data points randomly in the group $G_{PP}$ and data points do not change in other groups. Discriminatory dataset 2 (DA 2) is generated by sampling 3,000 data points randomly in the group $G_{PP}$ and data points do not change in other groups.
In each discriminatory dataset, we sample 2,000 data points as the discriminatory test dataset and the remaining as the training dataset.
\subsubsection{Synthetic Data Tested with Discriminatory and Fair Datasets}

\begin{table}[]
	\scalebox{0.92}{
	\begin{tabular}{lcclclclclclcl}
		\hline
		&            & \multicolumn{10}{c}{Test with discriminatory test dataset}                                                                                                    \\ \hline
		& Method     & \multicolumn{2}{c}{Acc}    & \multicolumn{2}{c}{Dis}    & \multicolumn{2}{c}{GPP} & \multicolumn{2}{c}{GUP} & \multicolumn{2}{c}{GPN} & \multicolumn{2}{c}{GUN} \\ \hline
		\multirow{4}{*}{LR}  & DA 1 (ORI) & \multicolumn{2}{c}{0.8815} & \multicolumn{2}{c}{0.2705} & \multicolumn{2}{c}{183} & \multicolumn{2}{c}{586} & \multicolumn{2}{c}{626} & \multicolumn{2}{c}{605} \\ 
		& DA 2 (ORI) & \multicolumn{2}{c}{0.8875} & \multicolumn{2}{c}{0.3642} & \multicolumn{2}{c}{104} & \multicolumn{2}{c}{628} & \multicolumn{2}{c}{642} & \multicolumn{2}{c}{626} \\ 
		& DA 1 (FS)  & \multicolumn{2}{c}{0.8825} & \multicolumn{2}{c}{0.2076} & \multicolumn{2}{c}{232} & \multicolumn{2}{c}{537} & \multicolumn{2}{c}{627} & \multicolumn{2}{c}{604} \\ 
		& DA 2 (FS)  & \multicolumn{2}{c}{0.8730} & \multicolumn{2}{c}{0.2890} & \multicolumn{2}{c}{159} & \multicolumn{2}{c}{573} & \multicolumn{2}{c}{642} & \multicolumn{2}{c}{626} \\ \hline
		\multirow{4}{*}{SVM} & DA 1 (ORI) & \multicolumn{2}{c}{0.8825} & \multicolumn{2}{c}{0.2664} & \multicolumn{2}{c}{188} & \multicolumn{2}{c}{581} & \multicolumn{2}{c}{629} & \multicolumn{2}{c}{602} \\ 
		& DA2 (ORI)  & \multicolumn{2}{c}{0.8880} & \multicolumn{2}{c}{0.3724} & \multicolumn{2}{c}{102} & \multicolumn{2}{c}{630} & \multicolumn{2}{c}{649} & \multicolumn{2}{c}{619} \\ 
		& DA 1 (FS)  & \multicolumn{2}{c}{0.8825} & \multicolumn{2}{c}{0.2097} & \multicolumn{2}{c}{231} & \multicolumn{2}{c}{538} & \multicolumn{2}{c}{628} & \multicolumn{2}{c}{603} \\ 
		& DA 2 (FS)  & \multicolumn{2}{c}{0.8745} & \multicolumn{2}{c}{0.3130} & \multicolumn{2}{c}{149} & \multicolumn{2}{c}{583} & \multicolumn{2}{c}{655} & \multicolumn{2}{c}{476} \\ \hline
	\end{tabular}
}
\caption{Two discriminatory datasets tested on the discriminatory test dataset in ORI method and the proposed fairness-enhanced method (FS)  with LR and SVM. We show accuracy (Acc), discrimination level (Dis) and the number of data points of each group in the discriminatory test dataset after classification.}
\end{table}

Table 1 shows that our method is able to reduce discrimination level when training datasets have different discrimination levels. For example, more data points are classified into the Protected group with positive labels $G_{PP}$ after implementing our method, and discrimination level of DA 1 reduces from 0.2705 to 0.2076 in LR. It is also note that accuracy does not decrease much with the proposed FS method. For example, accuracy of DA 2 reduces from 0.8825 to 0.8730 in LR. 
\begin{table}[]
	\scalebox{0.92}{
		\begin{tabular}{lcclclclclclcl}
			\hline
			&            & \multicolumn{10}{c}{Test with fair test   dataset}                                                                                                              \\ \hline
			& Method     & \multicolumn{2}{c}{Acc}    & \multicolumn{2}{c}{Dis}    & \multicolumn{2}{c}{GPP} & \multicolumn{2}{c}{GUP} & \multicolumn{2}{c}{GPN} & \multicolumn{2}{c}{GUN} \\ \hline
			\multirow{4}{*}{LR}  & DA1 (ORI)  & \multicolumn{2}{c}{0.8701} & \multicolumn{2}{c}{0.0484} & \multicolumn{2}{c}{438} & \multicolumn{2}{c}{556} & \multicolumn{2}{c}{492} & \multicolumn{2}{c}{514} \\ 
			& DA 2 (ORI) & \multicolumn{2}{c}{0.8535} & \multicolumn{2}{c}{0.1018} & \multicolumn{2}{c}{376} & \multicolumn{2}{c}{618} & \multicolumn{2}{c}{483} & \multicolumn{2}{c}{523} \\ 
			& DA 1 (FS)  & \multicolumn{2}{c}{0.8790} & \multicolumn{2}{c}{0.0161} & \multicolumn{2}{c}{474} & \multicolumn{2}{c}{520} & \multicolumn{2}{c}{496} & \multicolumn{2}{c}{510} \\ 
			& DA 2 (FS)  & \multicolumn{2}{c}{0.8810} & \multicolumn{2}{c}{0.0062} & \multicolumn{2}{c}{471} & \multicolumn{2}{c}{523} & \multicolumn{2}{c}{483} & \multicolumn{2}{c}{523} \\ \hline
			\multirow{4}{*}{SVM} & DA1 (ORI)  & \multicolumn{2}{c}{0.8700} & \multicolumn{2}{c}{0.0483} & \multicolumn{2}{c}{441} & \multicolumn{2}{c}{553} & \multicolumn{2}{c}{495} & \multicolumn{2}{c}{511} \\ 
			& DA 2 (ORI) & \multicolumn{2}{c}{0.8525} & \multicolumn{2}{c}{0.1118} & \multicolumn{2}{c}{372} & \multicolumn{2}{c}{622} & \multicolumn{2}{c}{489} & \multicolumn{2}{c}{517} \\ 
			& DA1 (FS)   & \multicolumn{2}{c}{0.8790} & \multicolumn{2}{c}{0.0168} & \multicolumn{2}{c}{474} & \multicolumn{2}{c}{520} & \multicolumn{2}{c}{496} & \multicolumn{2}{c}{510} \\ 
			& DA 2 (FS)  & \multicolumn{2}{c}{0.8775} & \multicolumn{2}{c}{0.0272} & \multicolumn{2}{c}{460} & \multicolumn{2}{c}{534} & \multicolumn{2}{c}{493} & \multicolumn{2}{c}{513} \\ \hline
		\end{tabular}%
	}
	\caption{Two discriminatory datasets tested on the fair test dataset in ORI method and the proposed fairness-enhanced method (FS) with LR and SVM. We show accuracy (Acc), discrimination level (Dis) and the number of data points of each group in the fair test dataset after classification.}
	\label{tab:my-table}
\end{table}

We test the biased datasets with the proposed FS method on the fair test dataset with LR and SVM, and results are shown in Table 2. 
With the proposed FS method, discrimination level decreases and accuracy increases. More specifically, discrimination level decreases from 0.1018 to 0.0062 and accuracy increases from 0.8535 to 0.8810 in the DA 2.
Discrimination level with the discriminatory test dataset is much higher than with the fair test dataset. We attribute this to the evaluation bias. Discriminatory dataset and discriminatory test data have the same data distribution, and thus the size of each group in the discriminatory test dataset is not equal. Even if the trained classifier is fair, the result may still be unfair. In real-world datasets, test datasets are sampled from the whole datasets and thus can contain evaluation bias.
\subsection{Discussion and Summery}
\subsubsection{Discussion}

We discuss on how the proposed FS framework is able to reduce discrimination in terms of discrimination decomposition into discrimination in bias, variance and noise. 
Discrimination in bias depends on the model choice. As we observe in the experiments, very broadly, LR can achieve a lower discrimination level than SVM. 
Discrimination in variance relates to the training data. Unlabeled data help to reduce discrimination in variance by increasing the size of training data. Ensemble learning helps to reduce discrimination in variance by averaging the training results from base models. 	
An appropriate unlabeled data size, sample size and ensemble size in our framework is able to help reduce more discrimination in variance. 
Discrimination in noise depends on the quality of data. Training with unlabeled data may bring discrimination in noise. However, ensemble learning offsets this effect.
When the same model is used, the benefit of unlabeled data in discrimination reduction depends on the impact of unlabeled data on discrimination in variance and discrimination in noise.

\subsubsection{Summary}
From these experiments, we see that the FS framework is able to reduce representation discrimination with a better trade-off between accuracy and discrimination.
In the proposed FS framework, discrimination reduction in variance is usually more than the discrimination incurred by label noise.
However, all the factors in the framework – model choice, unlabeled data size, ensemble size, sample size – each make their own particular contribution to increasing accuracy while ensuring fair representation.


%
\section{Related Work}
In recent years, much research on fair machine learning has been undertaken. The following subsections summarize the three main streams of this work.
\subsection{Pre-processing Methods}
Pre-processing methods eliminate the discrimination by adjusting the training data by ways of suppression, reweighing or sampling to obtain fair datasets before training \cite{calders2009building,kamiran2012data,calmon2017optimized}. Also, learning fair intermediate representations in the pre-process phase has received much attention.  
\cite{zemel2013learning} was the first to open up fair machine learning by learning fair intermediate representations. The basic idea is that mapping the training data to a transformed space where as much useful information as possible is retained, but the dependencies between sensitive attributes and class labels are removed.
Many researchers have subsequently studied fair representation learning with different methods, such as adversary learning \cite{madras2018learning,pmlr-v89-song19a,feng2019learning,zhang2018mitigating,mcnamara2019costs}. These methods are based on using a classifier to predict sensitive attributes as adversarial components.
The advantage of pre-precessing methods is that these methods can apply to all algorithms and tasks. Note that pre-processing approaches cannot be employed to eliminate discrimination arising from the algorithm itself. 
\subsection{In-processing Methods}
In-processing methods avoid discrimination with fair constraints
\cite{kamishima2012fairness} used regularizer term to penalize discrimination to enforce non-discrimination in the learning objective.  
\cite{zafar2017fairness,pmlr-v54-zafar17a,donini2018empirical} designed fairness constraints to achieve fair classification, where the fairness constraint is enforced by weakening the correlation between sensitive attribute and labels.
In \cite{agarwal2018reductions,JMLRCotter,DBLP:journals/corr/abs-1804-06500}, the constrained optimization problem is formulated as a two-player game and fairness definitions are formalized as linear inequalities.
Other recent work have a similar spirit to enforce fairness by adding constraints to the objective \cite{komiyama2018nonconvex,aghaei2019learning}.
The advantage of in-processing methods is that the level of fairness and accuracy can be controlled by the threshold of fairness constraints. 
However, fairness constraints are often irregular and need to be relaxed for optimization, and thus the solution may not be convergent.
\subsection{Post-processing Methods}
A third approach to achieving fairness is post-processing, where a learned classifier is modified to adjust the decisions to be non-discriminatory for different groups.
\cite{hardt2016equality} proposed an approach to use of post-processing to ensure fairness criteria of equal opportunity and equal odds and subsequent work include \cite{kim2019multiaccuracy,lohia2019bias} However, it is not guaranteed to find the most accurate fair classifier \cite{woodworth2017learning}, and requires test-time access to the protected attribute, which might not be available.

\subsection{Comparison with other work}
Existing fair methods focus on supervised and unsupervised learning, and these methods cannot be applied to SSL directly. As far as we know, only \cite{NIPS2019_9437,noroozi2019leveraging} considered fair SSL. In \cite{NIPS2019_9437},  data is used to learn the output conditional probability, and unlabeled data is used for calibration in the post-processing phase. This method is to eliminate the aggregation discrimination, while the proposed FS method is to reduce representation discrimination. In \cite{noroozi2019leveraging}, the proposed method is built on neural networks for SSL in the in-processing phase, and this method is to reduce measurement discrimination.
In \cite{kamiran2012data}, representation discrimination is reduced by uniform sampling and preferential sampling, while in some cases not enough data in minority group can be sampled to generate a fair dataset. Our work make use of unlabeled data to form fairer datasets and theoretically analyze the discrimination via decomposition in bias, variance and noise.
In our paper, we study the fair SSL based on label and unlabeled data in the pre-processing phase and our goal is to use labeled data to reduce representation discrimination, and in turn achieve a better trade-off between accuracy and discrimination.

\section{Conclusion and Future Work}

In future work, we intend to explore designs for fairness constraints that make use of unlabeled data to enforce fairness in the in-processing phase. 
Further, we have an assumption in this paper that labeled and unlabeled have the same distribution. However, this assumption may not hold in some real-world cases.
Hence, another research direction is to how to achieve fair semi-supervised learning where labeled and unlabeled data have different data distributions. 

\appendices

\section*{Acknowledgment}
This work is supported by an ARC Discovery Project (DP190100981) from Australian Research Council, Australia; and in part by NSF under grants III-1526499, III-1763325, III-1909323, and CNS-1930941. 
\ifCLASSOPTIONcaptionsoff
  \newpage
\fi

\bibliographystyle{IEEEtran}
\bibliography{TKDE}

\begin{thebibliography}{10}
\providecommand{\url}[1]{#1}
\csname url@samestyle\endcsname
\providecommand{\newblock}{\relax}
\providecommand{\bibinfo}[2]{#2}
\providecommand{\BIBentrySTDinterwordspacing}{\spaceskip=0pt\relax}
\providecommand{\BIBentryALTinterwordstretchfactor}{4}
\providecommand{\BIBentryALTinterwordspacing}{\spaceskip=\fontdimen2\font plus
\BIBentryALTinterwordstretchfactor\fontdimen3\font minus
  \fontdimen4\font\relax}
\providecommand{\BIBforeignlanguage}[2]{{%
\expandafter\ifx\csname l@#1\endcsname\relax
\typeout{** WARNING: IEEEtran.bst: No hyphenation pattern has been}%
\typeout{** loaded for the language `#1'. Using the pattern for}%
\typeout{** the default language instead.}%
\else
\language=\csname l@#1\endcsname
\fi
#2}}
\providecommand{\BIBdecl}{\relax}
\BIBdecl

\bibitem{chouldechova2017fair}
A.~Chouldechova, ``Fair prediction with disparate impact: A study of bias in
  recidivism prediction instruments,'' \emph{Big data}, vol.~5, no.~2, pp.
  153--163, 2017.

\bibitem{Obermeyer447}
\BIBentryALTinterwordspacing
Z.~Obermeyer, B.~Powers, C.~Vogeli, and S.~Mullainathan, ``Dissecting racial
  bias in an algorithm used to manage the health of populations,''
  \emph{Science}, vol. 366, no. 6464, pp. 447--453, 2019. [Online]. Available:
  \url{https://science.sciencemag.org/content/366/6464/447}
\BIBentrySTDinterwordspacing

\bibitem{calders2009building}
T.~Calders, F.~Kamiran, and M.~Pechenizkiy, ``Building classifiers with
  independency constraints,'' in \emph{2009 IEEE International Conference on
  Data Mining Workshops}.\hskip 1em plus 0.5em minus 0.4em\relax IEEE, 2009,
  pp. 13--18.

\bibitem{pmlr-v54-zafar17a}
M.~B. Zafar, I.~Valera, M.~G. Rodriguez, and K.~P. Gummadi, ``{Fairness
  Constraints: Mechanisms for Fair Classification},'' in \emph{Proceedings of
  the 20th International Conference on Artificial Intelligence and Statistics},
  vol.~54, 20--22 Apr 2017, pp. 962--970.

\bibitem{hardt2016equality}
M.~Hardt, E.~Price, N.~Srebro \emph{et~al.}, ``Equality of opportunity in
  supervised learning,'' in \emph{Advances in neural information processing
  systems}, 2016, pp. 3315--3323.

\bibitem{dwork2012fairness}
C.~Dwork, M.~Hardt, T.~Pitassi, O.~Reingold, and R.~Zemel, ``Fairness through
  awareness,'' in \emph{Proceedings of the 3rd innovations in theoretical
  computer science conference}.\hskip 1em plus 0.5em minus 0.4em\relax ACM,
  2012, pp. 214--226.

\bibitem{louizos2015variational}
C.~Louizos, K.~Swersky, Y.~Li, M.~Welling, and R.~Zemel, ``The variational fair
  autoencoder,'' \emph{arXiv preprint arXiv:1511.00830}, 2015.

\bibitem{jung2019eliciting}
C.~Jung, M.~Kearns, S.~Neel, A.~Roth, L.~Stapleton, and Z.~S. Wu, ``Eliciting
  and enforcing subjective individual fairness,'' \emph{arXiv preprint
  arXiv:1905.10660}, 2019.

\bibitem{kusner2017counterfactual}
M.~J. Kusner, J.~Loftus, C.~Russell, and R.~Silva, ``Counterfactual fairness,''
  in \emph{Advances in Neural Information Processing Systems}, 2017, pp.
  4066--4076.

\bibitem{kilbertus2017avoiding}
N.~Kilbertus, M.~R. Carulla, G.~Parascandolo, M.~Hardt, D.~Janzing, and
  B.~Sch{\"o}lkopf, ``Avoiding discrimination through causal reasoning,'' in
  \emph{Advances in Neural Information Processing Systems}, 2017, pp. 656--666.

\bibitem{kamiran2012data}
F.~Kamiran and T.~Calders, ``Data preprocessing techniques for classification
  without discrimination,'' \emph{Knowledge and Information Systems}, vol.~33,
  no.~1, pp. 1--33, 2012.

\bibitem{zemel2013learning}
R.~Zemel, Y.~Wu, K.~Swersky, T.~Pitassi, and C.~Dwork, ``Learning fair
  representations,'' in \emph{International Conference on Machine Learning},
  2013, pp. 325--333.

\bibitem{madras2018learning}
D.~Madras, E.~Creager, T.~Pitassi, and R.~Zemel, ``Learning adversarially fair
  and transferable representations,'' \emph{arXiv preprint arXiv:1802.06309},
  2018.

\bibitem{pmlr-v89-song19a}
J.~Song, P.~Kalluri, A.~Grover, S.~Zhao, and S.~Ermon, ``Learning controllable
  fair representations,'' in \emph{Proceedings of the 22nd International
  Conference on Artificial Intelligence and Statistics (AISTATS) 2019,},
  vol.~89, 16--18 Apr 2019, pp. 2164--2173.

\bibitem{kamishima2012fairness}
T.~Kamishima, S.~Akaho, H.~Asoh, and J.~Sakuma, ``Fairness-aware classifier
  with prejudice remover regularizer,'' in \emph{Joint European Conference on
  Machine Learning and Knowledge Discovery in Databases}.\hskip 1em plus 0.5em
  minus 0.4em\relax Springer, 2012, pp. 35--50.

\bibitem{pmlr-v97-kleindessner19b}
M.~Kleindessner, S.~Samadi, P.~Awasthi, and J.~Morgenstern, ``Guarantees for
  spectral clustering with fairness constraints,'' in \emph{Proceedings of the
  36th International Conference on Machine Learning}, vol.~97, Long Beach,
  California, USA, 09--15 Jun 2019, pp. 3458--3467.

\bibitem{NIPS2018_7613}
I.~Chen, F.~D. Johansson, and D.~Sontag, ``Why is my classifier
  discriminatory?'' in \emph{Advances in Neural Information Processing Systems
  31}, 2018, pp. 3539--3550.

\bibitem{suresh2019framework}
H.~Suresh and J.~V. Guttag, ``A framework for understanding unintended
  consequences of machine learning,'' \emph{arXiv preprint arXiv:1901.10002},
  2019.

\bibitem{domingos2000unified}
P.~Domingos, ``A unified bias-variance decomposition,'' in \emph{Proceedings of
  17th International Conference on Machine Learning}, 2000, pp. 231--238.

\bibitem{lee2013pseudo}
D.-H. Lee, ``Pseudo-label: The simple and efficient semi-supervised learning
  method for deep neural networks,'' in \emph{Workshop on Challenges in
  Representation Learning, ICML}, vol.~3, 2013, p.~2.

\bibitem{breiman1996bagging}
L.~Breiman, ``Bagging predictors,'' \emph{Machine learning}, vol.~24, no.~2,
  pp. 123--140, 1996.

\bibitem{zhu2005semi}
X.~J. Zhu, ``Semi-supervised learning literature survey,'' University of
  Wisconsin-Madison Department of Computer Sciences, Tech. Rep., 2005.

\bibitem{calmon2017optimized}
F.~Calmon, D.~Wei, B.~Vinzamuri, K.~N. Ramamurthy, and K.~R. Varshney,
  ``Optimized pre-processing for discrimination prevention,'' in \emph{Advances
  in Neural Information Processing Systems}, 2017, pp. 3992--4001.

\bibitem{feng2019learning}
R.~Feng, Y.~Yang, Y.~Lyu, C.~Tan, Y.~Sun, and C.~Wang, ``Learning fair
  representations via an adversarial framework,'' \emph{arXiv preprint
  arXiv:1904.13341}, 2019.

\bibitem{zhang2018mitigating}
B.~H. Zhang, B.~Lemoine, and M.~Mitchell, ``Mitigating unwanted biases with
  adversarial learning,'' in \emph{Proceedings of the 2018 AAAI/ACM Conference
  on AI, Ethics, and Society}, 2018, pp. 335--340.

\bibitem{mcnamara2019costs}
D.~McNamara, C.~S. Ong, and R.~C. Williamson, ``Costs and benefits of fair
  representation learning,'' in \emph{Proceedings of the 2019 AAAI/ACM
  Conference on AI, Ethics, and Society}, 2019, pp. 263--270.

\bibitem{zafar2017fairness}
M.~B. Zafar, I.~Valera, M.~Gomez~Rodriguez, and K.~P. Gummadi, ``Fairness
  beyond disparate treatment \& disparate impact: Learning classification
  without disparate mistreatment,'' in \emph{Proceedings of the 26th
  International Conference on World Wide Web}, 2017, pp. 1171--1180.

\bibitem{donini2018empirical}
M.~Donini, L.~Oneto, S.~Ben-David, J.~S. Shawe-Taylor, and M.~Pontil,
  ``Empirical risk minimization under fairness constraints,'' in \emph{Advances
  in Neural Information Processing Systems}, 2018, pp. 2791--2801.

\bibitem{agarwal2018reductions}
A.~Agarwal, A.~Beygelzimer, M.~Dud{\'\i}k, J.~Langford, and H.~Wallach, ``A
  reductions approach to fair classification,'' \emph{arXiv preprint
  arXiv:1803.02453}, 2018.

\bibitem{JMLRCotter}
A.~Cotter, H.~Jiang, M.~Gupta, S.~Wang, T.~Narayan, S.~You, and K.~Sridharan,
  ``Optimization with non-differentiable constraints with applications to
  fairness, recall, churn, and other goals,'' \emph{Journal of Machine Learning
  Research}, vol.~20, no. 172, pp. 1--59, 2019.

\bibitem{DBLP:journals/corr/abs-1804-06500}
A.~Cotter, H.~Jiang, and K.~Sridharan, ``Two-player games for efficient
  non-convex constrained optimization,'' \emph{CoRR}, 2018.

\bibitem{komiyama2018nonconvex}
J.~Komiyama, A.~Takeda, J.~Honda, and H.~Shimao, ``Nonconvex optimization for
  regression with fairness constraints,'' in \emph{International conference on
  machine learning}, 2018, pp. 2737--2746.

\bibitem{aghaei2019learning}
S.~Aghaei, M.~J. Azizi, and P.~Vayanos, ``Learning optimal and fair decision
  trees for non-discriminative decision-making,'' \emph{arXiv preprint
  arXiv:1903.10598}, 2019.

\bibitem{kim2019multiaccuracy}
M.~P. Kim, A.~Ghorbani, and J.~Zou, ``Multiaccuracy: Black-box post-processing
  for fairness in classification,'' in \emph{Proceedings of the 2019 AAAI/ACM
  Conference on AI, Ethics, and Society}, 2019, pp. 247--254.

\bibitem{lohia2019bias}
P.~K. Lohia, K.~N. Ramamurthy, M.~Bhide, D.~Saha, K.~R. Varshney, and R.~Puri,
  ``Bias mitigation post-processing for individual and group fairness,'' in
  \emph{Icassp 2019-2019 ieee international conference on acoustics, speech and
  signal processing (icassp)}.\hskip 1em plus 0.5em minus 0.4em\relax IEEE,
  2019, pp. 2847--2851.

\bibitem{woodworth2017learning}
B.~Woodworth, S.~Gunasekar, M.~I. Ohannessian, and N.~Srebro, ``Learning
  non-discriminatory predictors,'' in \emph{Conference on Learning Theory},
  2017, pp. 1920--1953.

\bibitem{NIPS2019_9437}
E.~Chzhen, C.~Denis, M.~Hebiri, L.~Oneto, and M.~Pontil, ``Leveraging labeled
  and unlabeled data for consistent fair binary classification,'' in
  \emph{Advances in Neural Information Processing Systems 32}, 2019, pp.
  12\,739--12\,750.

\bibitem{noroozi2019leveraging}
V.~Noroozi, S.~Bahaadini, S.~Sheikhi, N.~Mojab, and P.~S. Yu, ``Leveraging
  semi-supervised learning for fairness using neural networks,'' \emph{arXiv
  preprint arXiv:1912.13230}, 2019.

\end{thebibliography}

\begin{IEEEbiography}[{\includegraphics[width=1in,height=1.25in,clip,keepaspectratio]{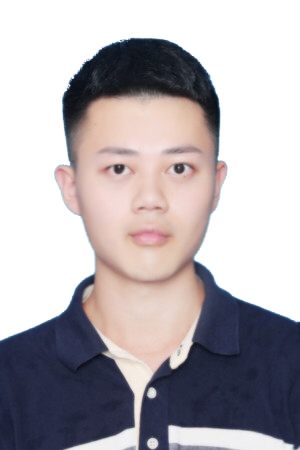}}]{Tao Zhang}
    works towards his Ph.D degree with the school of Computer Science in the University of Technology Sydney, Australia.
	His research interests include privacy preserving, algorithmic fairness, and machine learning.
\end{IEEEbiography}

\begin{IEEEbiography} [{\includegraphics[width=1in,height=1.25in,clip,keepaspectratio]{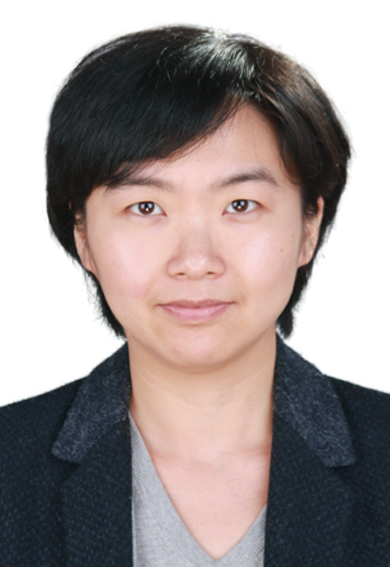}}]{Tianqing Zhu}
	 received the B.Eng. degree in chemistry and M.Eng. degree in automation from Wuhan University, Wuhan, China, in 2000 and 2004, respectively, and the Ph.D. degree in computer science from Deakin University, Geelong, Australia, in 2014. 
	 
	 She is currently a Associate Professor in the Faulty of Engineering and Information Technology with the School of Computer Science, University of Technology Sydney, Sydney, Australia. Before that, she was a Lecturer in the School of Information Technology, Deakin University, from 2014 to 2018. Her research interests include privacy preserving, data mining, and network security
\end{IEEEbiography}

\begin{IEEEbiography}[{\includegraphics[width=1in,height=1.25in,clip,keepaspectratio]{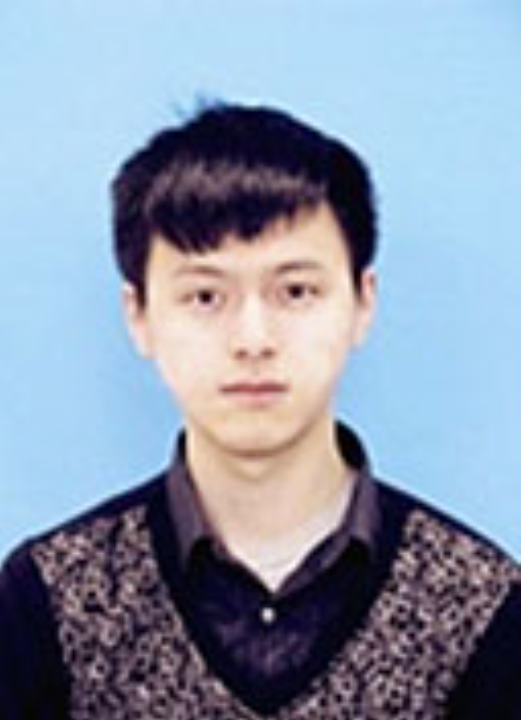}}]{Jing Li} 
	 received the B.Eng and M.Eng degrees in computer science and technology from Northwestern Polytechnical University, Xi’an, China, in 2015 and 2018, respectively.  
	 
	 Currently, he is pursuing the Ph.D degree with the Centre for Artificial Intelligence in the University of Technology Sydney, Australia. His research interests include machine learning and privacy preserving.

\end{IEEEbiography}

\begin{IEEEbiography}[{\includegraphics[width=1in,height=1.25in,clip,keepaspectratio]{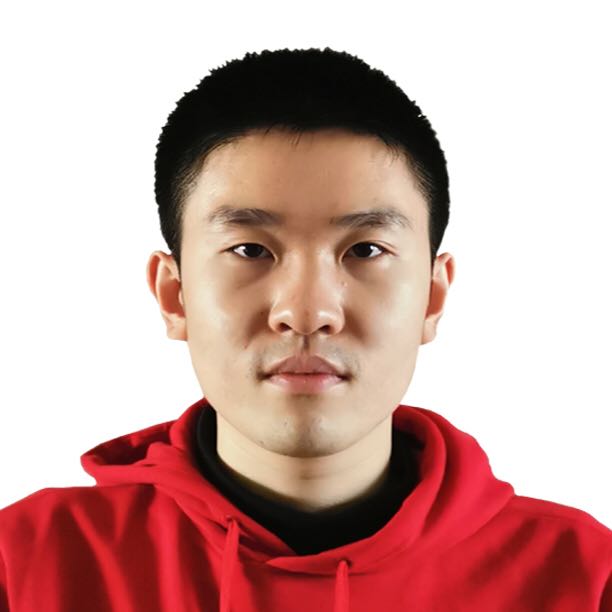}}]{Mengde Han}
	 is a PhD student at University of Technology Sydney with a focus on Local Differential Privacy. He completed his Master's at the Johns Hopkins University.
\end{IEEEbiography}

\begin{IEEEbiography}[{\includegraphics[width=1in,height=1.25in,clip,keepaspectratio]{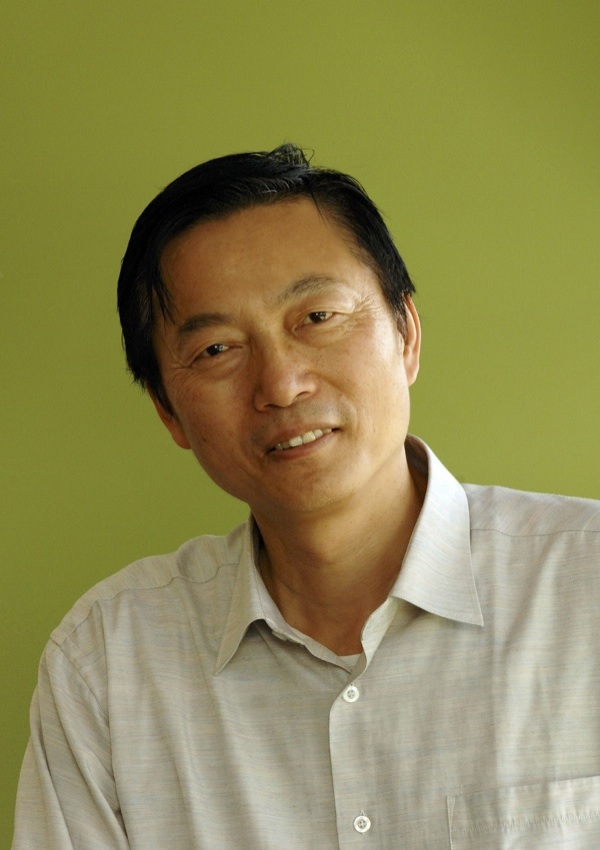}}]
	{Wanlei Zhou}
	received the B.Eng and M.Eng degrees from Harbin Institute of Technology, Harbin, China in 1982 and 1984, respectively, and the PhD degree from The Australian National University, Canberra, Australia, in 1991, all in Computer Science and Engineering. He also received a DSc degree (a higher Doctorate degree) from Deakin University in 2002. He is currently the Head of School of Computer Science in University of Technology Sydney (UTS). Before joining UTS, Professor Zhou held the positions of Alfred Deakin Professor, Chair of Information Technology, and Associate Dean (International Research Engagement) of Faculty of Science, Engineering and Built Environment, Deakin University. His research interests include security and privacy, bioinformatics, and e-learning. Professor Zhou has published more than 400 papers in refereed international journals and refereed international conferences proceedings, including many articles in IEEE transactions and journals.
\end{IEEEbiography}

\begin{IEEEbiography}[{\includegraphics[width=1in,height=1.25in,clip,keepaspectratio]{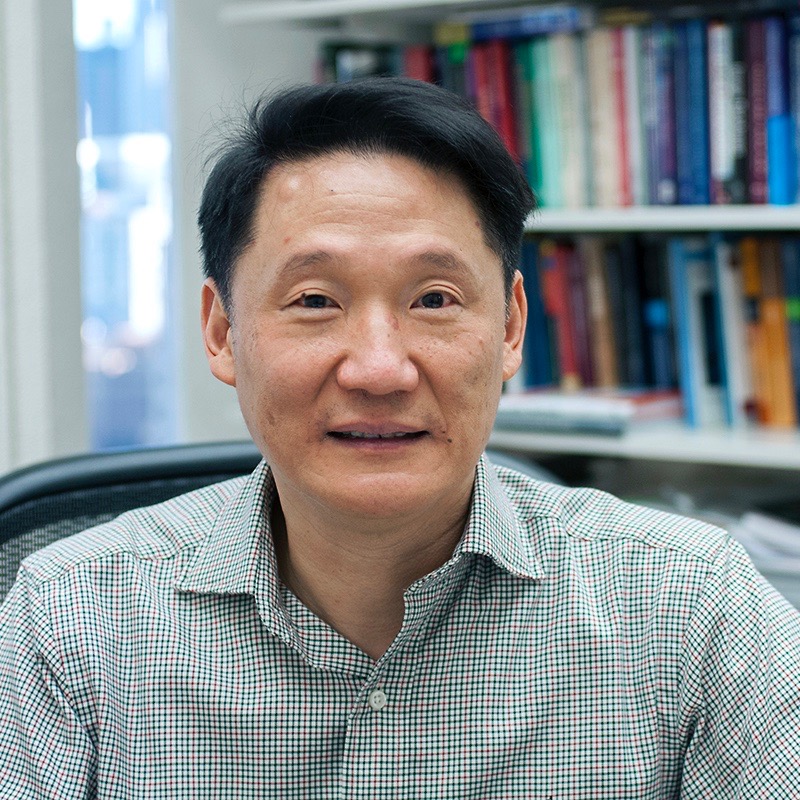}}]
	{Philip S. Yu}
	received the B.S. degree in electrical engineering from National Taiwan University, Taipei, Taiwan, the M.S. and Ph.D. degrees in electrical engineering from Stanford University, Stanford, CA, USA, and the M.B.A. degree from New York University, New York, NY, USA. He was with IBM, Armonk, NY, USA, where he was a Manager of the Software Tools and Techniques Department with the Thomas J. Watson Research Center. He is a Distinguished Professor of computer science with the University of Illinois at Chicago, Chicago, IL, USA, where he also holds the Wexler Chair in information technology. He has published over 1200 papers in peer-reviewed journals, such as the IEEE Transactions on Knowledge and Data Engineering, ACM Transactions on Knowledge Discovery from Data, VLDBJ, and the ACM Transactions on Intelligent Systems and Technology and conferences, such as KDD, ICDE, WWW, AAAI, SIGIR, ICML, and CIKM. He holds or has applied for over 300 U.S. patents. His current research interests include data mining, data streams, databases, and privacy. Dr. Yu was a recipient of the ACM SIGKDD 2016 Innovation Award for his influential research and scientific contributions on mining, fusion, and anonymization of Big Data and the IEEE Computer Society 2013 Technical Achievement Award. He was the Editor-in-Chief of ACM Transactions on Knowledge Discovery from Data (2011-2017) and IEEE Transactions on Knowledge and Data Engineering (2001-2004). He is a fellow of ACM.
	
\end{IEEEbiography}
\end{document}